\PassOptionsToPackage{table}{xcolor}
\documentclass[runningheads]{llncs}

\usepackage[preprint,year=2026]{eccv}

\newcommand{\ours}{\textbf{MolSight}}
\newcommand{\vi}{v_i}          
\newcommand{\gi}{g_i}          
\newcommand{\Lsig}{\mathcal{L}_{\text{SigLIP}}}
\newcommand{\Lhyb}{\mathcal{L}_{\text{Hybrid}}}

\newcommand{\best}[1]{\textbf{#1}}       
\newcommand{\second}[1]{\underline{#1}}  

\usepackage{eccvabbrv}

\usepackage{graphicx}
\usepackage{tabularx}
\usepackage{booktabs}
\usepackage{xcolor}
\usepackage{multirow}
\usepackage{subcaption}
\usepackage{amsmath,amssymb}
\usepackage{mathtools}
\usepackage{algorithm}
\usepackage{algpseudocode}
\usepackage{amsmath}
\usepackage{amssymb}
\usepackage[pagebackref, breaklinks, colorlinks]{hyperref}
\usepackage{makecell} 

\usepackage{hyperref}

\usepackage{orcidlink}

\begin{document}

\title{MolSight: Molecular Property Prediction with Images} 

\titlerunning{MolSight: Molecular Property Prediction with Images}

\author{
Aaditya Baranwal\inst{1} \and
Akshaj Gupta\inst{1,2} \and
Yogesh S Rawat\inst{1} \and
Shruti Vyas\inst{1}
}

\authorrunning{Aaditya Baranwal et al.}

\institute{University of Central Florida, Orlando, Florida, USA \and
Birla Institute of Technology and Science, Pilani, Rajasthan, India}

\maketitle

\begin{center}
  \includegraphics[width=\textwidth]{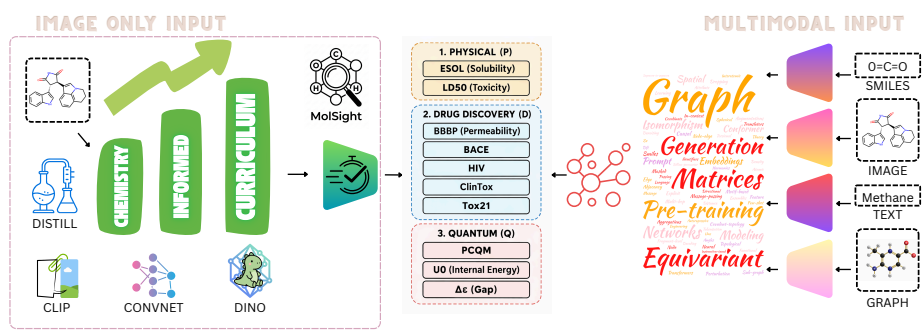}
  \captionof{figure}{
    \textbf{The \ours{} study: Architectures, strategies
    and downstream tasks.}
    \emph{Left:} \ours{} evaluates which visual architectures and
    pre-training strategies best encode chemical knowledge into a vision backbone, spanning ConvNets, self-supervised ViTs, and vision-language encoders.
    \emph{Center:} The resulting models are evaluated on 10 tasks across three molecular property categories using \emph{only} a 2D image at inference.
    \emph{Right:} Current MPP approaches require multi-modal inputs (graphs, conformers, LLMs) at inference, limiting throughput and accessibility.
  }
  \label{fig:study_overview}
\end{center}

\begin{abstract}
Every molecule ever synthesised can be drawn as a 2D skeletal diagram, yet in modern property prediction this universally available representation has received less focus in favour of molecular graphs, 3D conformers, or billion-parameter language models, each imposing its own computational and data-engineering overhead.
We present \textbf{MolSight}, the first systematic large-scale study of vision-based Molecular Property Prediction (MPP). Using 10 vision architectures, 7 pre-training strategies, and 2\,M molecule images, we evaluate performance across 10 downstream tasks
spanning physical-property regression, drug-discovery classification, and quantum-chemistry prediction.
To account for the wide variation in structural complexity across pre-training molecules, we further propose a \textbf{chemistry-informed curriculum}: five structural complexity descriptors partition the corpus into five tiers of increasing chemical difficulty, consistently outperforming non-curriculum baselines. We show that a single rendered bond-line image, processed by a vision encoder, is sufficient for competitive molecular property prediction, i.e. \emph{chemical insight from sight alone}. The best curriculum-trained configuration achieves the top result on \textbf{5 of 10} benchmarks and top two on \textbf{all 10}, at \textbf{\textit{80$\times$ lower}} FLOPs than the nearest multi-modal competitor.

\end{abstract}

\section{Introduction}
\label{sec:intro}

\begin{figure*}[t]
\centering
\includegraphics[width=\textwidth]{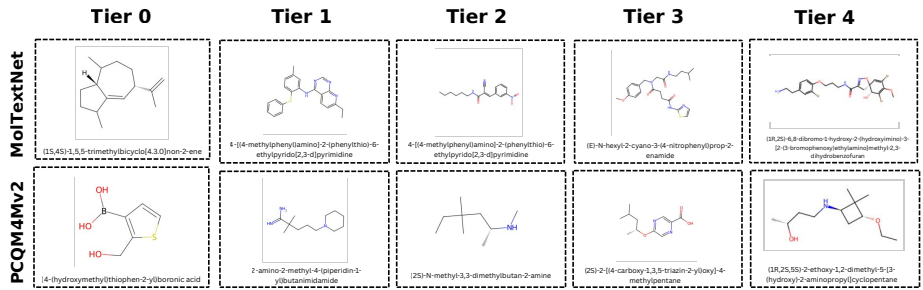}
\caption{
Example input samples from Tier 0 to Tier 4 from the Pretraining Datasets.
}
\label{fig:curriculum_tiers}
\end{figure*}

Since the 19th century,
chemists have communicated molecular structure through 2D skeletal diagrams, a visual language so expressive that an expert can infer reactivity, polarity, and rough pharmacological profiles at a glance.
Every compound in PubChem, ChEMBL, and ZINC can be rendered as such a diagram from a SMILES string in milliseconds, yet this universally available representation has been relatively underexplored
in modern molecular property prediction (MPP).
Instead, the field has developed several modality-specific approaches:
graph neural networks~\cite{gilmer2017neural,yang2019chemprop} that require explicit covalent-topology construction, 3D transformer models~\cite{zhou2022unimol} that demand expensive conformer generation, and instruction-tuned LLMs~\cite{edwards2022molt5} that invoke billion-parameter autoregressive decoding.
Each approach
delivers strong task-specific performance, but each also imposes its own data-engineering overhead and computational cost, creating a practical bottleneck for real-time screening at million-compound scales~\cite{adak2024molvision}.

\noindent In this work, we investigate whether this complexity is strictly necessary.
A 2D bond-line diagram encodes rich structural information: atomic identity, covalent connectivity, aromaticity, stereochemistry, and formal charge, while not capturing 3D conformation (Fig.~\ref{fig:attn_comparison}).
Our central hypothesis is: \emph{can a vision encoder, guided by the right pre-training strategy, learn to read these diagrams for competitive property prediction?}
We posit that the answer is yes, not by ignoring the value of 3D or graph-based approaches, but by showing that models pre-trained on natural images
and chemistry-informed scheduling (progressively ordering training examples by structural complexity)
achieve competitive results.

\noindent \ours{} tests this hypothesis at scale.
We systematically evaluate \textbf{10 vision architectures} under \textbf{7 pre-training strategies} on \textbf{2\,M molecule images} across \textbf{10 downstream tasks}---the largest study of vision-based MPP to date.
Across all settings, we observe that:
(i)~pre-training strategy matters far more than architecture choice; (ii)~a chemistry-informed curriculum that sequences molecules from simple hydrocarbons to complex macrocycles consistently outperforms random sampling; and (iii)~our best curriculum-trained model achieves the top result on \textbf{5 of 10 benchmarks} and ranks in the top two on \textbf{all 10}, at $80\times$ lower FLOPs than the nearest multi-modal competitor.

\paragraph{Contributions:}

\textbf{(1) Comprehensive pre-training study.} The first systematic large-scale evaluation of vision-based MPP: \textbf{10 architectures} (4~pure-vision, 6~vision-language) $\times$ \textbf{7 strategies} $\times$ \textbf{10 tasks}, yielding a detailed empirical map of which architecture--strategy combinations best encode chemical knowledge for each property category (Sec.~\ref{sec:study_design}).
\textbf{(2) Chemistry-informed curriculum.} Five structural complexity descriptors partition the pre-training corpus into five chemically grounded tiers: from pure hydrocarbons to polycyclic macrocycles, producing consistent gains across every architecture tested (Sec.~\ref{sec:curriculum}).
\textbf{(3) Competitive image-only inference.} Our best curriculum-trained model is competitive with or surpasses Uni-Mol~\cite{zhou2022unimol} (a leading 3D-conformer GNN)
on 8 of 10 benchmarks and MolVision~\cite{adak2024molvision} (a recent vision-language approach)
on 7 of 10, using only a rendered 2D image at \textbf{\textit{80$\times$ lower compute}} (Sec.~\ref{sec:experiments}).

\section{Related Work}
\label{sec:related}

\paragraph{Graph-based molecular representation learning:}
Plethora of works including Message Passing Neural Networks~\cite{gilmer2017neural} established the canonical graph-learning framework for molecules, with subsequent extensions improving multi-task generalisation (ChemProp~\cite{yang2019chemprop}), attention pooling (AttentiveFP~\cite{xiong2020attentivefp}), and long-range interactions (GPS~\cite{rampavsek2022recipe}) have relied on 3D graph information of molecules. Uni-Mol~\cite{zhou2022unimol,lu2023unimol} pushes accuracy further by pre-training a 3D transformer on millions of DFT conformers. While highly effective, these methods require explicit molecular topology, and 3D variants demand computationally expensive conformer generation thereby limiting their applicability in high-throughput settings. We study an image-only alternative that requires no molecular graph at inference.

\paragraph{Language model approaches:}
SMILES and SELFIES-based~\cite{krenn2022selfies} transformers, including ChemBERT~\cite{chithrananda2020chemberta}, GROVER~\cite{rong2020grover}, and MolT5~\cite{edwards2022molt5}, transfer large-scale masked-language pre-training to property prediction. These models benefit from vast text corpora, and leverage autoregressive abilities of transformers yet remain tethered to sequential tokenisation, which discards the 2D spatial layout explicitly embedded in skeletal diagrams. \ours{} explores whether this spatial information can be directly exploited by operating on rendered images.

\paragraph{Vision and vision-language approaches:}
Self-supervised contrastive learning on molecular images (SimCLR~\cite{chen2020simclr}, MoCo) reduces annotation burden but lacks cross-modal chemical grounding. CLIP~\cite{radford2021clip} and SigLIP~\cite{zhai2023siglip} provide image-text alignment; DINOv2~\cite{oquab2023dinov2} and EVA-CLIP~\cite{sun2023evaclip} push the accuracy--efficiency frontier; Long-CLIP~\cite{zhang2024longclip} extends the token window for chemical descriptions. MolVision~\cite{adak2024molvision} adapts LLaVA-style VLMs to predict properties from multi-modal prompts, while GPT-4o~\cite{openai2024gpt4o} and Janus-Pro~7B~\cite{chen2025januspro} rely on in-context learning. Recent benchmarks of LLMs on chemistry, such as ChemPro~\cite{baranwal2026chempro}, find that even frontier LLMs struggle on quantitative chemistry tasks. All of these approaches, however, retain multi-modal inputs or billion-parameter decoders at inference. \ours{} distills their representational strengths into a single vision encoder
that requires only an image at test/inference time to produce exceeding or competitive results.

\paragraph{Cross-modal knowledge distillation:}
Transferring representations from a teacher modality to a more accessible student has succeeded in audio-visual~\cite{aytar2016soundnet}, image-text~\cite{kim2025cosmos}, camera-radar~\cite{zhao2024crkd}, and broader cross-modal distillation settings~\cite{huo2024c2kd,ji2024vexkd}.
In molecular ML, existing works compress large GNNs into smaller GNNs; to our knowledge, none target vision encoders as distillation recipients. \ours{} introduces a hybrid distillation loss that transfers 3D GNN embedding structure into the representational space of a 2D image encoder.

Table~\ref{tab:taxonomy} (supplementary) situates \ours{} within this landscape, highlighting its architectural simplicity at inference compared to methods reliant on 3D geometry, symbolic strings, or LLM decoders.

\section{Methodology}
\label{sec:study_design}

\ours{} investigates whether a general-purpose vision encoder, guided by the right pre-training, can serve as a competitive alternative to modality-specific molecular property predictors.
We systematically vary three design axes: the \emph{visual architecture} (Sec.~\ref{sec:architectures}), the \emph{pre-training strategy} (Sec.~\ref{sec:strategies}), and the \emph{downstream evaluation protocol} (Sec.~\ref{sec:tasks}).
The chemistry-informed curriculum, a key contribution, is detailed in Sec.~\ref{sec:curriculum}.

\subsection{Visual Architectures}
\label{sec:architectures}

Our study spans ten visual architectures across two broad families: \textbf{Pure-vision encoders} (ResNet-18, ResNet-50, DINOv2, DINOv3) and \textbf{Vision-language encoders} (SigLIP, SigLIP2, OpenCLIP, Long-CLIP, MetaCLIP, EVA-CLIP) (Table~\ref{tab:architectures}). Pure-vision encoders establish baselines and quantify the marginal value of cross-modal pre-training over self-supervised image representations. Vision-language encoders leverage pre-existing cross-modal alignment priors from large-scale image-text pre-training. For downstream molecular property prediction, we use only the visual encoder paired with a lightweight prediction head.

\begin{table*}[t!]
\centering
\scriptsize
\caption{
  \textbf{Visual architectures.}
  VL-capable models include a text encoder used \emph{only during pre-training};
  it is discarded at inference. Params: encoder only.
}
\label{tab:architectures}
\setlength{\tabcolsep}{3.8pt}
\begin{tabular}{l l l r r c}
\toprule
\textbf{Architecture} & \textbf{Backbone} & \textbf{Type}
  & \textbf{Params} & \textbf{Patch/Stride} & \textbf{VL?} \\
\midrule
ResNet-18          & ResNet-18     & ConvNet    &  12M & $32\times$ stride & No  \\
ResNet-50          & ResNet-50     & ConvNet    &  25M & $32\times$ stride & No  \\
DINOv2             & ViT-B/14      & Self-sup.  &  86M & $14\times14$ px   & No  \\
DINOv3             & ViT-B/14      & Self-sup.  &  86M & $14\times14$ px   & No  \\
SigLIP             & ViT-B/16      & VL         &  86M & $16\times16$ px   & Yes \\
SigLIP2            & ViT-B/16      & VL         &  86M & $16\times16$ px   & Yes \\
OpenCLIP           & ViT-B/16      & VL         &  86M & $16\times16$ px   & Yes \\
Long-CLIP          & ViT-B/16      & VL (ext.)  &  86M & $16\times16$ px   & Yes \\
MetaCLIP           & ViT-B/16      & VL         &  86M & $16\times16$ px   & Yes \\
EVA-CLIP           & ViT-B/16      & VL         &  86M & $16\times16$ px   & Yes \\
\bottomrule
\end{tabular}%
\end{table*}

\paragraph{Image representation:}
Architectures receive molecules as standardised $224\times224$ RGB
bond-line diagrams rendered by RDKit with white background and default atom
colouring. Pixel values are normalised per-channel to $[-1,1]$.
ViT: process the image as a sequence of non-overlapping patches;
ConvNets: apply strided convolutions.
A global pooled representation (ViT \texttt{[CLS]} or ResNet
global-average-pool) is $\ell_2$-normalised for loss computation or
downstream prediction.

\subsection{Pre-training Datasets and Strategies}
\label{sec:strategies}

\begin{figure*}[t!]
  \centering
  \includegraphics[width=\textwidth]{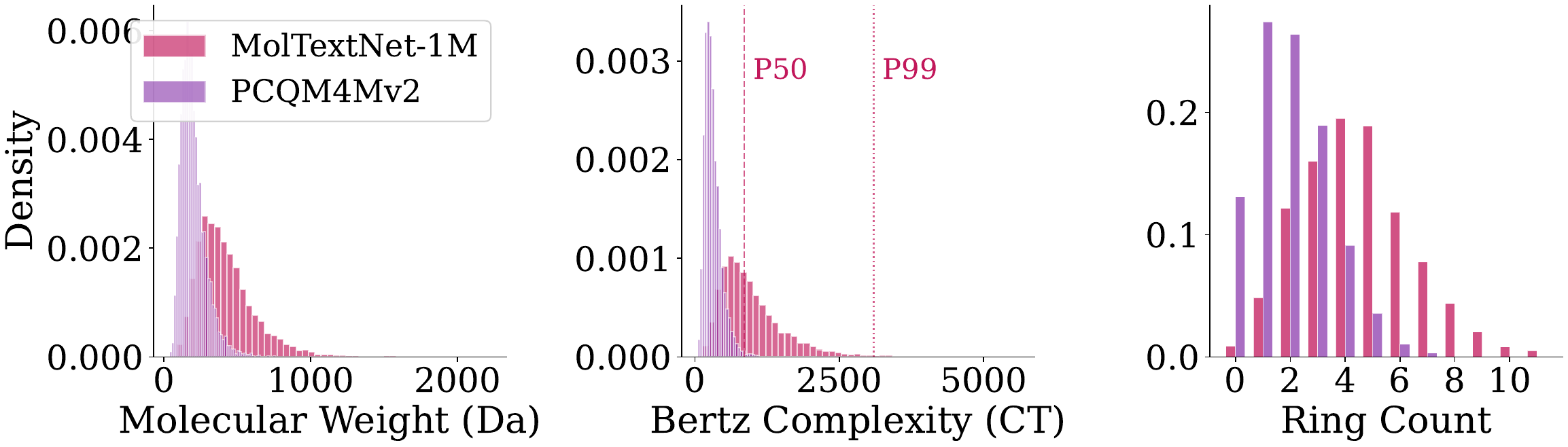}
  \caption{
    \textbf{Pre-training corpus descriptor distributions for MolTextNet-1M and PCQM4Mv2.}
    \emph{(a)}~Molecular weight; \emph{(b)}~Bertz topological complexity $CT$;
    \emph{(c)}~ring count.
    MolTextNet-1M molecules are ${\sim}2\times$ heavier and ${\sim}3\times$ more
    topologically complex than PCQM4Mv2, motivating a complexity-ordered curriculum.
  }
  \label{fig:datastats}
\end{figure*}

For pre-training, we use two complementary corpora totalling 2\,M molecule-image pairs (Fig.~\ref{fig:datastats}).
Together they cover distinct regions of chemical complexity, motivating a curriculum approach.

\paragraph{MolTextNet-1M} contains 1\,M drug-like molecules, each paired with a $224\times224$ RGB image and a SMILES string. Molecules in this corpus display high structural complexity (mean molecular weight $\text{MW}\approx 421$\,Da, mean Bertz complexity $CT\approx 998$), featuring dense ring systems and diverse functional groups spanning tens of thousands of unique Murcko scaffolds.

\paragraph{PCQM4Mv2} is the OGB quantum-chemistry dataset ($\sim$3.8\,M molecules with DFT-computed HOMO--LUMO gaps); we utilise the $\sim$1\,M subset with pre-computed molecular embeddings (described in Sec.~\ref{sec:strategies}).
In contrast to MolTextNet, PCQM4Mv2 consists of substantially simpler, low-molecular-weight structures (mean $\text{MW}\approx 202$\,Da, mean $CT \approx 313$) with few rings, making them well suited for grounding fundamental quantum-chemical properties.

We study seven pre-training strategies ranging from no chemical pre-training to curriculum-augmented GNN distillation
(Table~\ref{tab:strategies}).
All strategies share the same image rendering, normalisation, and downstream fine-tuning
protocol, isolating the effect of pre-training alone.

\begin{table*}[t!]
\centering
\caption{
  \textbf{Pre-training strategies.}
  Listed in order of increasing chemical supervision.
  $^*$: VL-capable models only.
  $^\dagger$: requires pre-computed Uni-Mol teacher embeddings.
  MolTN: MolTextNet.
}
\label{tab:strategies}
\setlength{\tabcolsep}{4pt}
\resizebox{\textwidth}{!}{%
\begin{tabular}{c l l l l c}
\toprule
\textbf{ID} & \textbf{Strategy} & \textbf{Key Objective} & \textbf{Modalities}
  & \textbf{Dataset}\\
\midrule
S1 & No Pre-training         & Supervised fine-tune only             & Image          & ---       \\
S2 & Image-Image (SimCLR)    & NT-Xent on augmented image pairs      & Image          & MolTN-1M \\
S3 & Image-SMILES (SigLIP)   & Sigmoid contrastive (Eq.~\ref{eq:siglip}) & Image + SMILES & MolTN-1M \\
S4 & Image-Description       & Sigmoid contrastive on NL descriptions & Image + Text & MolTN-1M \\
S5 & GNN Distillation        & Hybrid loss (Eq.~\ref{eq:hybrid})     & Image + GNN    & PCQM4Mv2$^\dagger$ \\
S6 & Curriculum       & S5 with tier-based scheduler (Sec.~\ref{sec:curriculum}) & Image + GNN & PCQM4Mv2$^\dagger$ \\
S7 & Joint SLIP              & S2 + S3 simultaneously ($\Lsig + \lambda\mathcal{L}_\text{SimCLR}$) & Image + SMILES & MolTN-1M \\
\bottomrule
\end{tabular}%
}
\end{table*}

\noindent \emph{S1 (Baseline):}
The vision encoder is initialised from weights pre-trained on natural images~\cite{deng2009imagenet} (ConvNets)
or from the original VL checkpoint (VL models) and fine-tuned directly on downstream tasks, providing a simple fine-tuning baseline.

\noindent \emph{S2 (Image-Image contrastive):}
Two augmented views of the same molecule image are
generated via random crops, flips, colour jitter, and
grayscale conversion.
The NT-Xent loss~\cite{chen2020simclr} maximises agreement between positive pairs:
\begin{equation}
  \mathcal{L}_\text{SimCLR} = -\sum_{i=1}^{N}
    \log \frac{\exp\!\bigl(\hat{v}_i^{(1)} \cdot \hat{v}_i^{(2)} / \tau\bigr)}
              {\sum_{k \ne i} \exp\!\bigl(\hat{v}_i^{(1)} \cdot \hat{v}_k^{(2)} / \tau\bigr)},
  \label{eq:simclr}
\end{equation}
where $\hat{v}_i^{(1)}$ and $\hat{v}_i^{(2)}$ are $\ell_2$-normalised embeddings of two augmented views of molecule $i$, and $\tau{=}0.07$ is a temperature parameter.

\noindent \emph{S3 (Image-SMILES):}
For VL-capable architectures, molecular images are aligned with canonical
SMILES strings via the SigLIP sigmoid contrastive loss~\cite{zhai2023siglip}.
With signed labels $\ell_{ij} = 2\cdot\mathbf{1}[i{=}j]-1$:
\begin{equation}
\footnotesize
  \Lsig(f_\theta, g_\phi)
  = -\frac{1}{N^2} \sum_{i,j}
    \log \sigma\!\bigl(\ell_{ij}\,\hat{v}_i^\top\hat{t}_j + \ell_{ij}\,b\bigr),
  \label{eq:siglip}
\end{equation}
where $\hat{v}_i \in \mathbb{R}^d$ and $\hat{t}_j \in \mathbb{R}^d$ are $\ell_2$-normalised image and text embeddings respectively, and $b$ is a learnable scalar bias.
Unlike CLIP's softmax, SigLIP treats each pair independently, enabling stable training at large batch sizes.

\noindent \emph{S4 (Image-Description):}
SMILES strings are replaced with natural-language molecular descriptions, including functional group enumerations and IUPAC.

\noindent \emph{S5 (GNN distillation):}
A frozen Uni-Mol~\cite{zhou2022unimol} teacher produces embeddings $\gi \in \mathbb{R}^{512}$ from 3D conformers, computed once offline.
A trainable projection $\phi$ maps teacher embeddings into the encoder's space, and a regression head $\mathcal{H}$ predicts the HOMO-LUMO gap $y_i$ (both discarded at inference).
The hybrid distillation loss is:
\begin{equation}
\footnotesize
  \Lhyb = \Lsig\!\bigl(\hat{v}_i, \hat{\tilde{g}}_i\bigr)
         + \alpha \bigl\|\phi(\gi) - \vi\bigr\|_2^2
         + \beta  \bigl\|\mathcal{H}(\vi) - y_i\bigr\|_2^2,
  \label{eq:hybrid}
\end{equation}
where $\hat{\tilde{g}}_i = \phi(\gi)/\|\phi(\gi)\|_2$ is the projected and normalised teacher embedding, $\vi$ is the image encoder output, $\mathcal{H}$ is a linear regression head discarded at inference, and $\alpha{=}10$, $\beta{=}1$ are loss weights.
The contrastive term aligns embedding manifolds; the MSE term anchors numerical proximity; the energy term encourages predictive structure in the representations.

\noindent \emph{S6 (Curriculum):}
Strategy S5 augmented with the chemistry-informed five-tier curriculum
scheduler described in Sec.~\ref{sec:curriculum}, evaluated on all
architectures.

\noindent \emph{S7 (Joint SLIP):}
S2 (SimCLR) and S3 (SigLIP SMILES alignment) are combined simultaneously: $\mathcal{L}_\text{SLIP} = \Lsig + \lambda\,\mathcal{L}_\text{SimCLR}$ with $\lambda = 1.0$. The text encoder is permanently discarded after pre-training.

\subsection{Downstream Evaluation Tasks}
\label{sec:tasks}

\begin{table}[b]
\centering
\scriptsize
\caption{
  \textbf{Downstream evaluation tasks.}
  P = physical property; D = drug discovery; Q = quantum chemistry.
  ROC-AUC (higher is better) for classification; MAE (lower is better) for
  regression.
  All scaffold splits follow MoleculeNet conventions.`}
\label{tab:tasks}
\setlength{\tabcolsep}{3.8pt}
\begin{tabular}{l l r l l c}
\toprule
\textbf{Task} & \textbf{Dataset} & \textbf{Size} & \textbf{Split}
  & \textbf{Metric} & \textbf{Cat.} \\
\midrule
ESOL              & Delaney~\cite{delaney2004esol}    & 1,128   & Scaffold & RMSE (log mol/L) & P \\
LD50              & Zhu et al.~\cite{zhu2009ld50}     & 7,413   & Random   & MAE (log)        & P \\
\midrule
BBBP              & MoleculeNet~\cite{wu2018moleculenet} & 2,039 & Scaffold & ROC-AUC        & D \\
BACE              & MoleculeNet~\cite{wu2018moleculenet}      & 1,513   & Scaffold & ROC-AUC         & D \\
ClinTox           & MoleculeNet~\cite{wu2018moleculenet}     & 1,491   & Scaffold & ROC-AUC         & D \\
Tox21             & MoleculeNet~\cite{wu2018moleculenet}    & 7,831   & Scaffold & ROC-AUC         & D \\
HIV               & MoleculeNet~\cite{wu2018moleculenet}   & 41,127  & Scaffold & ROC-AUC         & D \\
\midrule
QM9 ($U_0$)       & QM9~\cite{ramakrishnan2014quantum} & 133,885 & Random  & MAE (eV)        & Q \\
QM9 ($\Delta\varepsilon$) & QM9 ~\cite{ramakrishnan2014quantum}  & 133,885 & Random  & MAE (meV)       & Q \\
PCQM4Mv2          & OGB-LSC                            & 3.8M    & Official & MAE (eV)       & Q \\
\bottomrule
\end{tabular}%
\end{table}

Ten downstream tasks span three molecular property categories
(Table~\ref{tab:tasks}):

\noindent \textbf{Physical-property regression} (2 tasks):
  ESOL~\cite{delaney2004esol} (aqueous solubility, 1128 molecules) and
  LD50~\cite{zhu2009ld50} (acute oral toxicity, ${\sim}$7.4k molecules).
  Dominated by polarity, hydrogen bonding, and surface area that are all visible
  in 2D diagrams.

\noindent \textbf{Drug-discovery classification} (5 tasks): BBBP, BACE, ClinTox,
  Tox21, and HIV from MoleculeNet~\cite{wu2018moleculenet} with scaffold
  splits and ROC-AUC as the primary metric.
  These require recognition of functional-group patterns associated with
  permeability, binding, metabolic stability, and toxicophore identity.

\noindent \textbf{Quantum-chemistry prediction} (3 tasks):
  QM9~\cite{ramakrishnan2014quantum} atomisation energy ($U_0$) and HOMO-LUMO
  gap ($\Delta\varepsilon$) on the standard 110k/10k/10k random split; and
  PCQM4Mv2 HOMO-LUMO gap on the official OGB split.
  These are most strongly coupled to 3D electronic structure and represent
  the most challenging regime for a 2D vision encoder.

\noindent \textbf{Implementation Details}\\
\emph{Optimiser and schedule:} All pre-training uses Adam ($\text{lr}=10^{-4}$, $\beta_1=0.9$, $\beta_2=0.98$) for 10 epochs with cosine annealing and 10\% linear warmup. For DDP runs (GNN distillation, curriculum), gradient accumulation achieves an effective batch of 4096 with \texttt{no\_sync} for memory efficiency. \emph{Downstream fine-tuning:} A three-layer MLP head ($d \to 256 \to 128 \to 1$ for regression; sigmoid output for classification) is appended to the encoder's pooled representation. The final two transformer blocks (or last two convolutional stages for ResNets) are unfrozen and fine-tuned jointly at $10\times$ lower learning rate. All downstream experiments run for 50 epochs. ROC-AUC is reported for classification; MAE for regression.

\section{Chemistry-Informed Curriculum Learning}
\label{sec:curriculum}
\begin{figure}[t!]
  \centering
  \includegraphics[width=\textwidth]{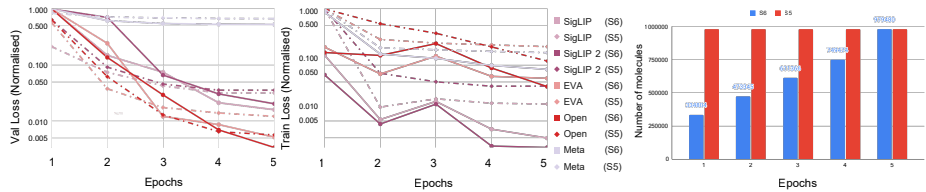}
  \caption{
    \textbf{Pre-training convergence with and without curriculum scheduling.}
    Left: validation loss for curriculum training (S6) vs random sampling (S5).
    Middle: training loss for S6 vs S5.
    Right: number of molecules processed per epoch under each strategy.
    Solid lines denote S6; dashed lines denote S5. Curriculum training yields smoother optimization and lower validation loss.
 }
  \label{fig:training_curves}
\end{figure}

An aspirin molecule (21 atoms, one aromatic ring) and a taxol molecule (113 atoms, four fused rings, three stereocentres) present vastly different visual complexity, yet random mini-batching treats them identically.
Curriculum learning~\cite{bengio2009curriculum} addresses this by sequencing examples from simple to complex.
We design a chemistry-informed curriculum that partitions the pre-training corpus using five structural descriptors, each targeting a distinct source of visual difficulty in molecular diagrams.

\noindent \textbf{Dataset Characterisation:}
Fig.~\ref{fig:datastats} profiles the pre-training corpora along the
structural dimensions most relevant to visual complexity.
MolTextNet molecules are ${\sim}2\times$ heavier (mean MW 421.2~Da vs.\
202.2~Da) and ${\sim}3\times$ more topologically complex
($\overline{CT} = 997.8$ vs.\ 313.1) than PCQM4Mv2, confirming that the
two corpora cover genuinely complementary regions of chemical space.
Within MolTextNet-1M, the 99th percentile of Bertz complexity
($CT = 3326$) is ${\sim}3.5\times$ the 50th percentile ($CT = 912$), extreme
non-uniformity that makes random sampling highly inefficient.

\noindent \textbf{Five Structural Complexity Descriptors} are computed for every pre-training molecule using
RDKit and a custom SMARTS-pattern library (31 functional-group patterns).
Together they cover the principal axes of visual complexity in 2D diagrams (full definitions in the supplementary):
\textbf{(1)~Scaffold Decoration} $D_\text{scaf}$: ratio of non-scaffold to total heavy atoms, capturing peripheral clutter;
\textbf{(2)~Functional-Group Rarity} $R$: mean inverse corpus prevalence of a molecule's functional groups (Eq.~\ref{eq:rarity});
\textbf{(3)~Conjugation Extent} $C$: size of the largest connected $\pi$-system;
\textbf{(4)~Aromatic Substitution Complexity} $S$: count and diversity of ring substitution patterns;
\textbf{(5)~Bertz Topological Complexity} $CT$~\cite{bertz1981first}: Shannon entropy of the bond-environment distribution (Eq.~\ref{eq:bertz}).
\begin{equation}
\footnotesize
  R(m) = \begin{cases}
    0 & \text{if } |F_m| = 0, \\[2pt]
    \frac{1}{|F_m|} \sum_{f \in F_m} \bigl(1 - P(f)\bigr)
      & \text{otherwise,}
  \end{cases}
  \label{eq:rarity}
\end{equation}
\begin{equation}
\footnotesize
  CT(m) = \tfrac{1}{2}\!\left[\Bigl(\sum_k n_k \log_2 n_k\Bigr)
    + n_e \log_2 n_e\right],
  \label{eq:bertz}
\end{equation}
where $F_m$ is the set of unique functional groups, $P(f)$ their corpus prevalence, and $n_e$, $\{n_k\}$ are the bond-environment count distribution.

\begin{figure*}[t!]
  \centering
  \footnotesize
  \includegraphics[width=\linewidth]{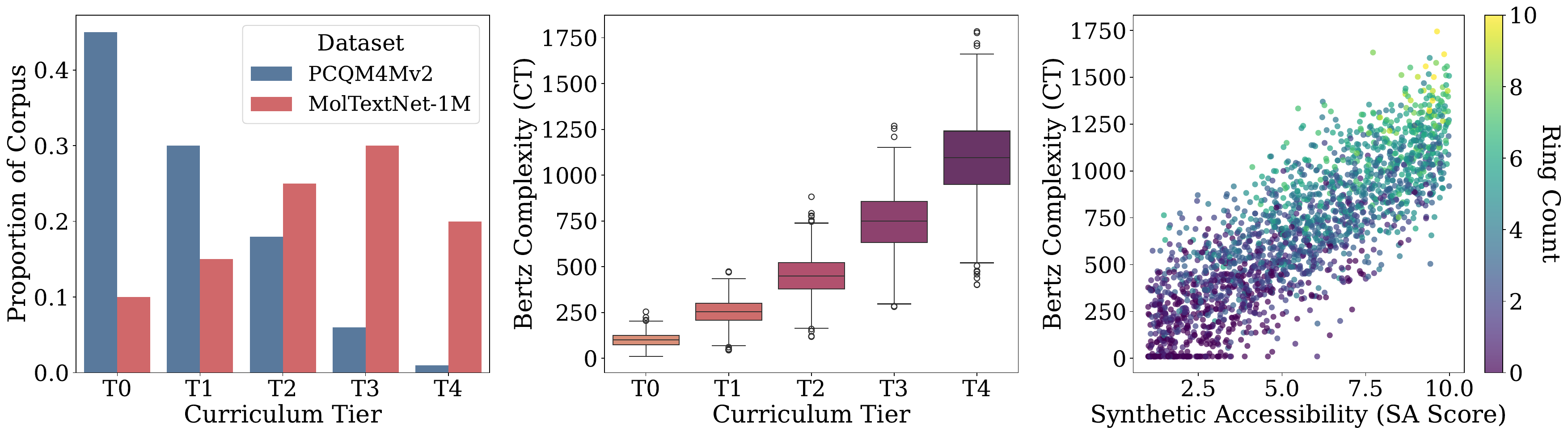}
  \caption{
    \textbf{Curriculum tier (CT) characterisation:}
    \emph{(a)}~Tier distribution (molecule counts) for MolTextNet-1M and PCQM4Mv2.
    PCQM4Mv2 is dominated by Tiers~0--2 (simple quantum-chemistry molecules);
    MolTextNet harbours a richer Tier~3--4 population (complex drug-like
    structures).
    \emph{(b)}~Boxplots of Bertz complexity $CT$ per tier for MolTextNet-1M,
    confirming monotone, well-separated complexity increase across tiers.
    \emph{(c)}~$CT$ vs.\ synthetic accessibility (SA score) for PCQM4Mv2,
    coloured by ring count; Tier~4 molecules occupy the high-complexity,
    low-accessibility corner.
  }
  \label{fig:curriculum}
\end{figure*}

\noindent \textbf{Tier Assignment and Training Schedule}
The five descriptors assign each molecule to one of five tiers via a deterministic rule set (Table~\ref{tab:curriculum_tiers}): Tier~0 represents low-complexity structures (e.g., pure hydrocarbons), while Tier~4 represents high-complexity, drug-like structures with rare functional groups, dense stereocenters, and macrocyclic scaffolds. As illustrated in Fig.~\ref{fig:curriculum}, the assignments effectively stratify the corpora, isolating the hardest examples (Tier 4) while confirming a monotonically increasing structural difficulty across the sequence.

\begin{table*}[b]

\centering
\caption{
  \textbf{Chemical tiers for curriculum scheduling.}
  Each tier is defined by a deterministic combination of the five descriptors. FG: Functional Group
  (Eqs.~\ref{eq:rarity}--\ref{eq:bertz}).
}
\label{tab:curriculum_tiers}
\begin{tabular}{clp{7cm}c}
\toprule
\textbf{Tier} & \textbf{Name} & \textbf{Defining Constraints}
  & \textbf{Stereo?} \\
\midrule
$T_0$ & Scaffolds       & $n_\text{het}=0$ (pure hydrocarbons)                        & No  \\
$T_1$ & Single FG       & $n_\text{FG}\le 2$; all FGs in top-6 most common            & No  \\
$T_2$ & Multi-FG        & $3 \le n_\text{FG} \le 5$; $S(m) \le 4$                    & Rarely \\
$T_3$ & Positional      & $S(m) > 4$ or ($CT/n_\text{ha} > 50$ and $n_\text{ring}\ge3$) & Sometimes \\
$T_4$ & Full Complexity & $n_\text{sc}>0$ or high $R(m)$; polycyclic, macrocycles   & Yes \\
\bottomrule
\end{tabular}%
\end{table*}

\noindent \textbf{Progressive scheduling:}
Training initiates at epoch $e{=}0$ utilizing only Tier $T_0$ molecules. At epoch $e \geq 1$, the active pool progressively expands: $\mathcal{D}_e = \bigcup_{k=0}^{\min(e,\,4)} T_k$. This progression allows the encoder to assimilate simpler bonding patterns prior to encountering extreme structural variants, ensuring stable representation learning over the varied corpus.

\section{Experiments and Analysis}
\label{sec:experiments}

We evaluate 10 architectures under each applicable strategy (S1--S6, Table~\ref{tab:strategies}) and fine-tune on each downstream task (Table~\ref{tab:tasks}), with all results averaged over three random seeds (see supplementary for per-seed statistics).

\subsection{Strategy Study: Which Pre-Training Strategy Matters Most?}
\label{sec:strategy_study}

\begin{figure*}[t!]
  \centering
  \begin{minipage}{0.52\linewidth}
    \centering
  \includegraphics[width=\textwidth]{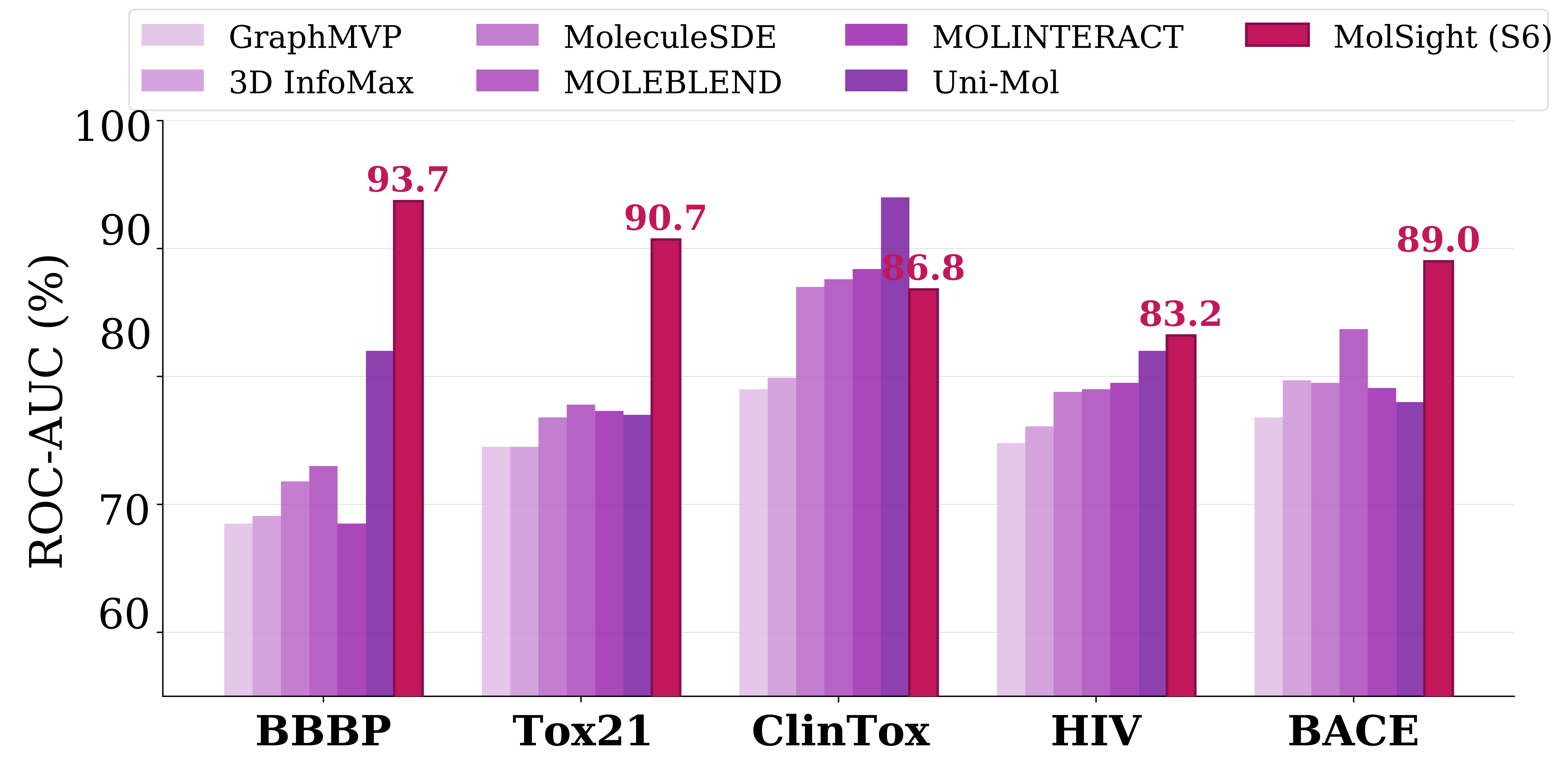}
  \caption{
    \textbf{Classification performance vs.\ GNN pre-training baselines (ROC-AUC \%).}
    \ours{} (deep rose) achieves the highest AUC on 4 of 5 tasks, competitive with graph-based methods that operate on explicit molecular topology.
    Uni-Mol$^\dagger$ (3D conformers) leads only on ClinTox.
  }
  \label{fig:classification_bars}
  \end{minipage}
  \hfill
  \begin{minipage}{0.42\linewidth}
    \centering
  \centering
  \includegraphics[width=0.95\textwidth]{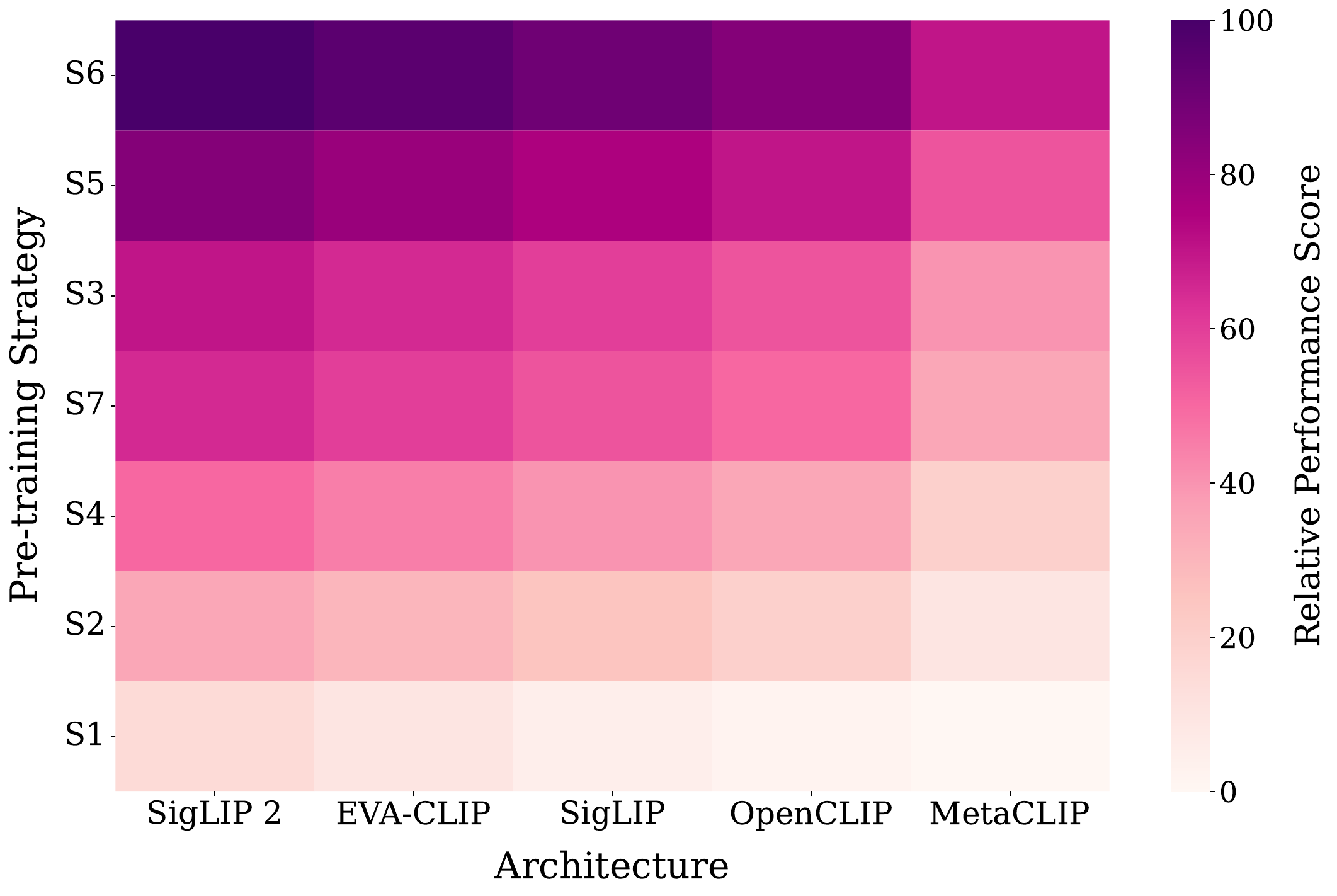}
  \caption{
    \textbf{Strategy $\times$ architecture performance.}
    Curriculum-augmented GNN distillation (S6) outperforms every other strategy regardless of backbone.
    The ordering S6${\succ}$S5${\succ}$S3/S7${\succ}$S4${\succ}$S2${\succ}$S1 is consistent across all tasks.
  }
  \label{fig:strategy_heatmap}
  \end{minipage}
\end{figure*}

Table~\ref{tab:strategy_results} reports performance across representative
(strategy, architecture) configurations on all available tasks. A clear hierarchy emerges (Fig.~\ref{fig:strategy_heatmap}).
\textbf{Curriculum (S6) ranks first on every task}, with substantial gains on quantum benchmarks: SigLIP2 under S6 achieves QM9-$\Delta\varepsilon$ MAE of 0.00697, a $31.7\%$ reduction over S3 (0.0102), while PCQM MAE drops from 0.549 to 0.095, an $82.7\%$ reduction once GNN distillation and curriculum scheduling incorporate the Uni-Mol teacher's 3D inductive bias.
Cross-modal strategies (S3, S4, S6) consistently outperform self-supervised
Image-Image contrastive learning (S2). Moving from S2 to S3 on SigLIP2 improves ESOL from 1.330 to 0.703 ($-47.1\%$), BBBP from 87.0\% to 93.6\%, and HIV from 74.0\% to 81.3\%. These results suggest that domain-specific supervision signals are essential: self-supervised visual pre-training alone does not provide sufficient chemical grounding.
\begin{table*}[b]
\centering
\caption{
  \textbf{Pre-training strategy comparison.}
  Each row is a representative (strategy, architecture) configuration.
  ESOL: RMSE ($\downarrow$); LD50: MAE ($\downarrow$);
  QM9 targets: MAE ($\downarrow$); PCQM: MAE ($\downarrow$);
  drug-discovery tasks: ROC-AUC ($\uparrow$, \%).
  \best{Bold}: best per column.
  \second{Underline}: second best.
  $^\dagger$: PCQM not applicable (no Uni-Mol teacher embeddings).
}
\label{tab:strategy_results}
\setlength{\tabcolsep}{3pt}
\resizebox{\textwidth}{!}{%
\begin{tabular}{cl l cc cc c ccccc}
\toprule
& & & \multicolumn{2}{c}{\textbf{Physical ($\downarrow$)}}
  & \multicolumn{2}{c}{\textbf{Quantum ($\downarrow$)}}
  &
  & \multicolumn{5}{c}{\textbf{Drug Discovery (AUC$\uparrow$, \%)}} \\
\cmidrule(lr){4-5}\cmidrule(lr){6-7}\cmidrule(lr){9-13}
\textbf{ID} & \textbf{Strategy} & \textbf{Architecture}
  & ESOL & LD50
  & QM9-$\Delta\varepsilon$ & PCQM
  &
  & BBBP & BACE & HIV & ClinTox & Tox21 \\
\midrule
S2 & Image-Image   & ResNet-50
  & 1.330 & 0.580 & 0.0100 & 7.635$^\dagger$ && 87.0 & 87.0 & 74.0 & 76.0 & 82.0 \\
S2 & Image-Image   & DINOv3
  & 0.774 & 0.630 & 0.0100 & 5.220$^\dagger$ && 89.5 & \best{89.5} & 74.3 & 80.3 & 81.0 \\
S3 & Image-SMILES  & SigLIP2
  & 0.703 & 0.403 & 0.0102 & 0.549 && \second{93.6} & 88.5 & \second{81.3} & 81.0 & \second{89.2} \\
S3 & Image-SMILES  & EVA-CLIP
  & 0.752 & 0.460 & 0.0112 & 0.288 && 91.7 & 84.0 & 78.0 & 80.0 & 86.0 \\
S4 & Image-Desc.   & Long-CLIP
  & \best{0.670} & 0.460 & 0.0122 & 0.416 && 91.0 & 84.0 & 77.0 & 83.0 & 85.0 \\
\midrule
S6 & \textbf{Curriculum} & \textbf{SigLIP2}
  & \second{0.627} & \best{0.369} & \best{0.00697} & \second{0.095}
  && \best{93.7} & \second{89.0} & \best{83.2} & \best{86.8} & \best{90.0} \\
S6 & \textbf{Curriculum} & \textbf{EVA-CLIP}
  & 0.717 & \second{0.384} & \second{0.0076} & \best{0.081}
  && 92.0 & 87.8 & 79.0 & \second{84.0} & \best{90.0} \\
\bottomrule
\end{tabular}%
}
\end{table*}

\noindent \emph{Image-SMILES vs.\ image-description:}
S3 (Image-SMILES) outperforms S4 (Image-Description) on quantum tasks
(QM9-$\Delta\varepsilon$: 0.0102 vs.\ 0.0122) and drug discovery (BBBP:
93.6\% vs.\ 91.0\%), though S4 achieves a marginally better ESOL (0.670
vs.\ 0.703).
Canonical SMILES uniquely encode the full molecular graph in compact form;
natural-language descriptions are partial and coarse.

\noindent \emph{Architecture sensitivity:}
Within S6, SigLIP2 and EVA-CLIP show complementary strengths: SigLIP2 leads
on QM9-$\Delta\varepsilon$ (0.00697) and drug-discovery tasks,
while EVA-CLIP achieves the best PCQM MAE (0.081).
A full per-architecture breakdown is provided in the supplementary.
Fig.~\ref{fig:strategy_ladder} illustrates this progressive improvement on three representative tasks, showing that each additional layer of chemical supervision: from coarse image-SMILES alignment (S3) to fine-grained GNN distillation (S5) to curriculum pacing (S6), yields measurable gains.

\begin{figure*}[t!]
  \centering
  \includegraphics[width=0.9\textwidth]{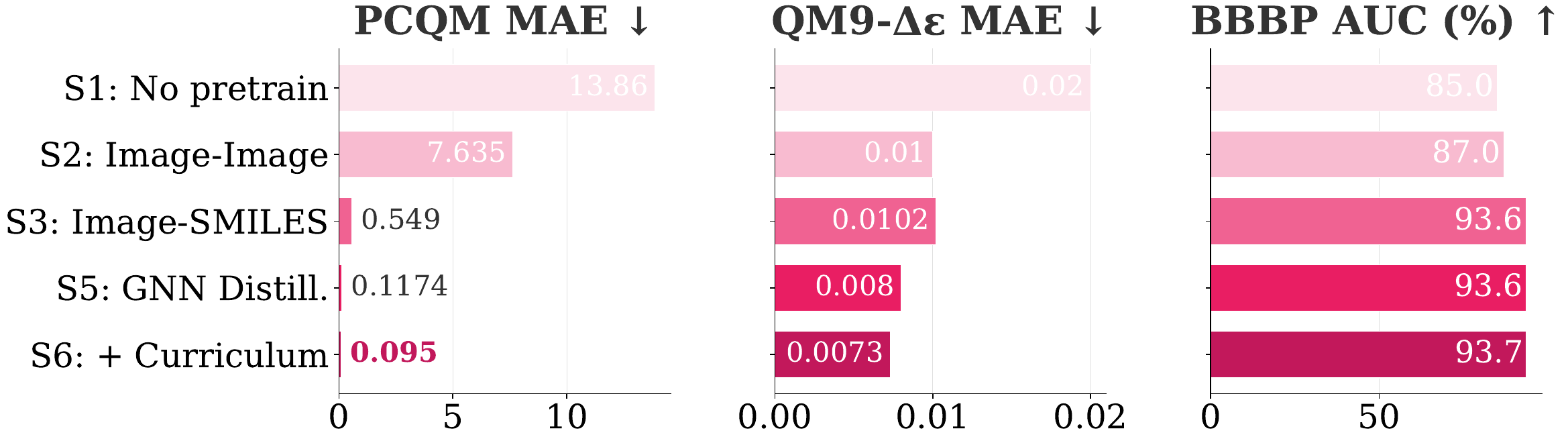}
  \caption{
    \textbf{Progressive improvement across pre-training strategies (SigLIP2).}
    From left to right: PCQM MAE, QM9-$\Delta\varepsilon$ MAE, and BBBP AUC.
    Each step up the supervision ladder yields measurable gains, with curriculum-augmented
    GNN distillation (S6) achieving the strongest performance on all three representative tasks.
  }
  \label{fig:strategy_ladder}
\end{figure*}

\subsection{Curriculum Impact and Ablation}
\label{sec:ablation_curriculum}

Table~\ref{tab:curriculum_ablation} isolates the curriculum's contribution
within the GNN distillation framework.
\begin{table}[b]
\centering
\caption{
  \textbf{Curriculum scheduling ablation (SigLIP2), decomposed.}
  S3 = Image-SMILES; S5 = GNN distillation (no curriculum); S6 = + curriculum. Three-seed mean.
  Bottom block isolates the distillation (S3$\to$S5) and curriculum (S5$\to$S6) contributions.
}
\label{tab:curriculum_ablation}
\setlength{\tabcolsep}{4pt}
\begin{tabular}{l c c c c c}
\toprule
\textbf{Config}
  & ESOL$\downarrow$ & LD50$\downarrow$
  & QM9-$\Delta\varepsilon$$\downarrow$ & PCQM$\downarrow$
  & BBBP$\uparrow$ \\
\midrule
S3: Image-SMILES         & 0.703 & 0.403 & 0.0102 & 0.549 & 93.6 \\
S5: GNN Distil., no curr.\ & 0.702 & 0.371 & 0.008 & 0.1174 & 93.6 \\
\textbf{S6: + Curriculum (Ours)} & \best{0.627} & \best{0.369} & \best{0.00697} & \best{0.095} & \best{93.7} \\
\midrule
$\Delta$ S3$\to$S5 (distillation)  & $-0.1\%$  & $-7.9\%$  & $-21.6\%$ & $-78.6\%$ & $+0.0$ \\
$\Delta$ S5$\to$S6 (curriculum)    & $-10.7\%$ & $-0.5\%$  & $-12.9\%$ & $-19.1\%$ & $+0.1$ \\
$\Delta$ S3$\to$S6 (combined)      & $-10.8\%$ & $-8.4\%$  & $-31.7\%$ & $-82.7\%$ & $+0.1$ \\
\bottomrule
\end{tabular}%
\end{table}

The curriculum provides consistent gains on quantum tasks: PCQM MAE drops from
0.549 (S3) to 0.095 (S6, $-82.7\%$), and QM9-$\Delta\varepsilon$ MAE
decreases by $31.7\%$ (0.0102 to 0.00697).
Physical-property prediction also benefits (ESOL $-10.8\%$, LD50 $-8.4\%$),
while drug-discovery AUC is largely unchanged, consistent with image-SMILES
alignment already capturing the functional-group signals central to
pharmacological classification.

\noindent \textbf{Empirical Validation}
Fig.~\ref{fig:training_curves} confirms that curriculum scheduling converges faster and reaches lower final loss than uniform random sampling across all tested architectures, with the largest benefit in early epochs where restricting exposure to simple tiers prevents destabilising gradients from complex molecules.

\subsection{State-of-the-Art Comparison}
\label{sec:sota}

\begin{table*}[b]
\centering
\caption{
  \textbf{State-of-the-art comparison.}
  \ours{}-Best: best curriculum-trained architecture per task.
  \ours{} (SigLIP2): single-model results.
  Both use only a 2D image at inference.
  \best{Bold}: best overall. \second{Underline}: second best.
  $\dagger$: requires 3D conformer generation.
  $\ddagger$: requires LLM/VLM at inference.
  Drug-discovery: Accuracy@0.5, matching the MolVision~\cite{adak2024molvision} VLM protocol; ROC-AUC for the same runs in Tab.~\ref{tab:strategy_results}.
} 
\label{tab:sota}
\setlength{\tabcolsep}{3pt}
\resizebox{\textwidth}{!}{%
\begin{tabular}{l l cc ccccc ccc}
\toprule
\multirow{2}{*}{\textbf{Category}}
  & \multirow{2}{*}{\textbf{Method}}
  & \multicolumn{2}{c}{\textbf{Physical ($\downarrow$)}}
  & \multicolumn{5}{c}{\textbf{Drug Discovery (Accuracy$\uparrow$, \%)}}
  & \multicolumn{3}{c}{\textbf{Quantum ($\downarrow$)}} \\
\cmidrule(lr){3-4}\cmidrule(lr){5-9}\cmidrule(lr){10-12}
  & & ESOL & LD50 & BBBP & BACE & ClinTox & Tox21 & HIV
  & QM9-$\Delta\varepsilon$ & QM9-$U_0$ & PCQM \\
\midrule
3D GNN
  & Uni-Mol$^\dagger$
  & 0.788 & 0.580
  & 82.0 & 78.0 & \best{94.0} & 77.0 & 82.0
  & \best{0.0047} & -- & \best{0.070} \\
3D GNN
  & Uni-Mol2$^\dagger$
  & -- & --
  & -- & -- & -- & -- & --
  & \best{0.0035} & -- & -- \\
LLM
  & GPT-4o (ICL)$^\ddagger$
  & 0.980 & 0.870
  & 77.0 & 56.0 & 59.0 & 42.0 & 82.0
  & 8.380 & 17.943 & 0.680 \\
VLM
  & Janus-Pro 7B (ICL)$^\ddagger$
  & \second{0.610} & 0.720
  & 68.0 & 78.0 & 83.0 & 69.0 & \second{92.0}
  & 8.530 & \second{17.211} & 0.620 \\
VLM
  & BLIP-2 (LoRA)$^\ddagger$
  & 1.070 & \second{0.490}
  & \best{93.0} & \best{86.0} & \second{89.0} & \best{99.0} & \second{92.0}
  & 4.920 & 22.120 & 1.990 \\
\midrule
\multirow{2}{*}{\shortstack[l]{\textbf{Image-only}\\\textbf{(Ours)}}}
  & \ours{} (SigLIP2, S6)
  & 0.627 & \best{0.369}
  & \second{89.0} & \second{82.0} & \best{94.0} & \second{91.0} & \best{97.0}
  & \second{0.00697} & \best{1.99} & 0.095 \\
  & \ours{}-Best
  & \best{0.515} & \best{0.369}
  & \second{89.0} & \second{82.0} & \best{94.0} & \second{91.0} & \best{97.0}
  & \second{0.00697} & \best{1.99} & \second{0.081} \\
\bottomrule
\end{tabular}%
}
\end{table*}

Throughout this section, \ours{} (S6) refers to our best single-model curriculum-trained configuration (SigLIP2 under Strategy 6); \ours{}-Best selects the highest-performing architecture per task.

\begin{figure*}[t!]
  \centering
  \begin{minipage}{0.51\linewidth}
    \centering
    \includegraphics[width=\linewidth]{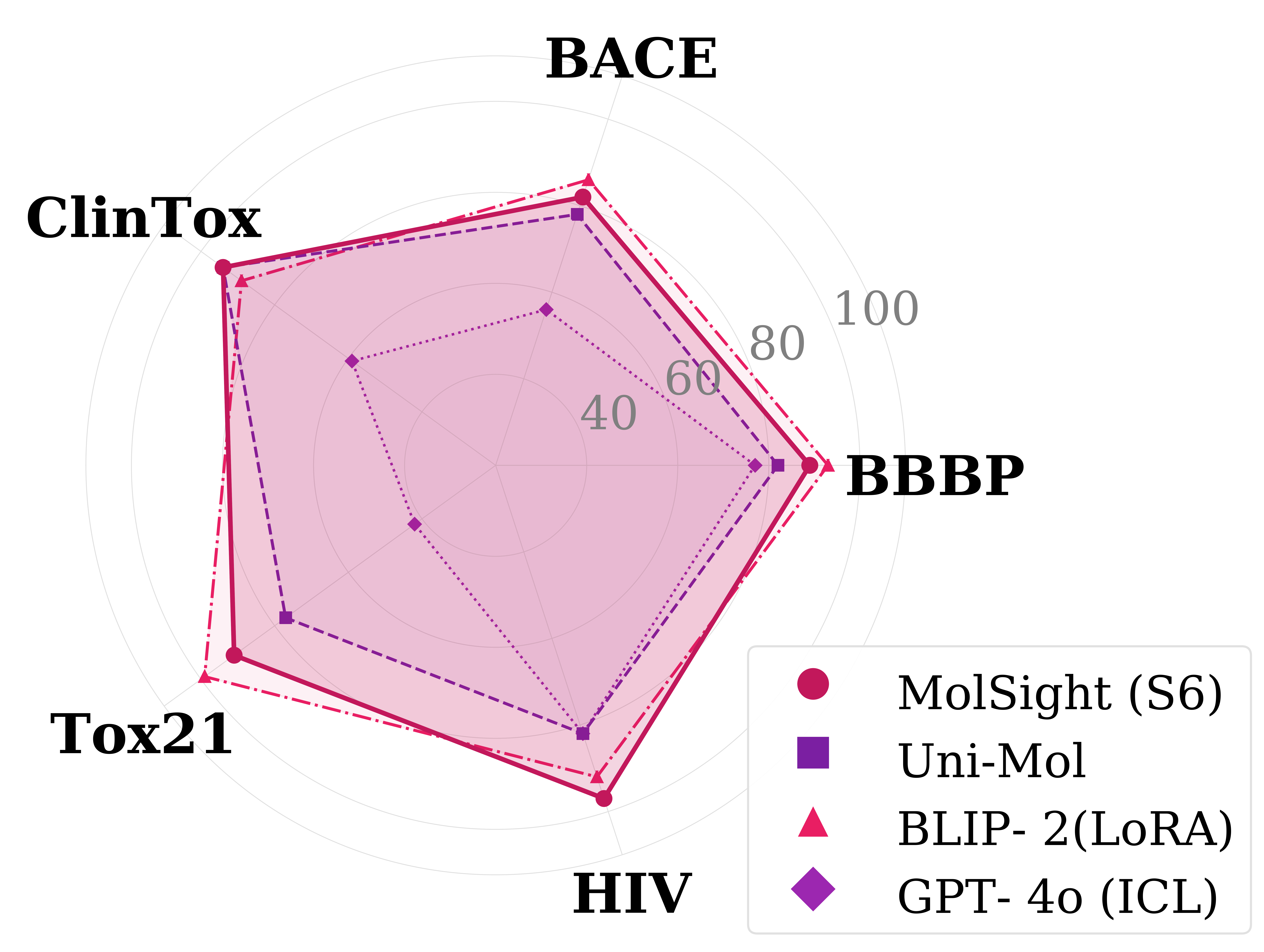}
    \caption{
      \textbf{Drug-discovery classification \%.}
      \ours{} (solid) achieves the largest and most balanced polygon, leading on Tox21, HIV, and BBBP.
      None dominates across all tasks: BLIP-2 leads on BACE but collapses on ClinTox; GPT-4o consistently scores low.
    }
    \label{fig:radar_classification}
  \end{minipage}
  \hfill
  \begin{minipage}{0.44\linewidth}
    \centering
    \includegraphics[width=\linewidth]{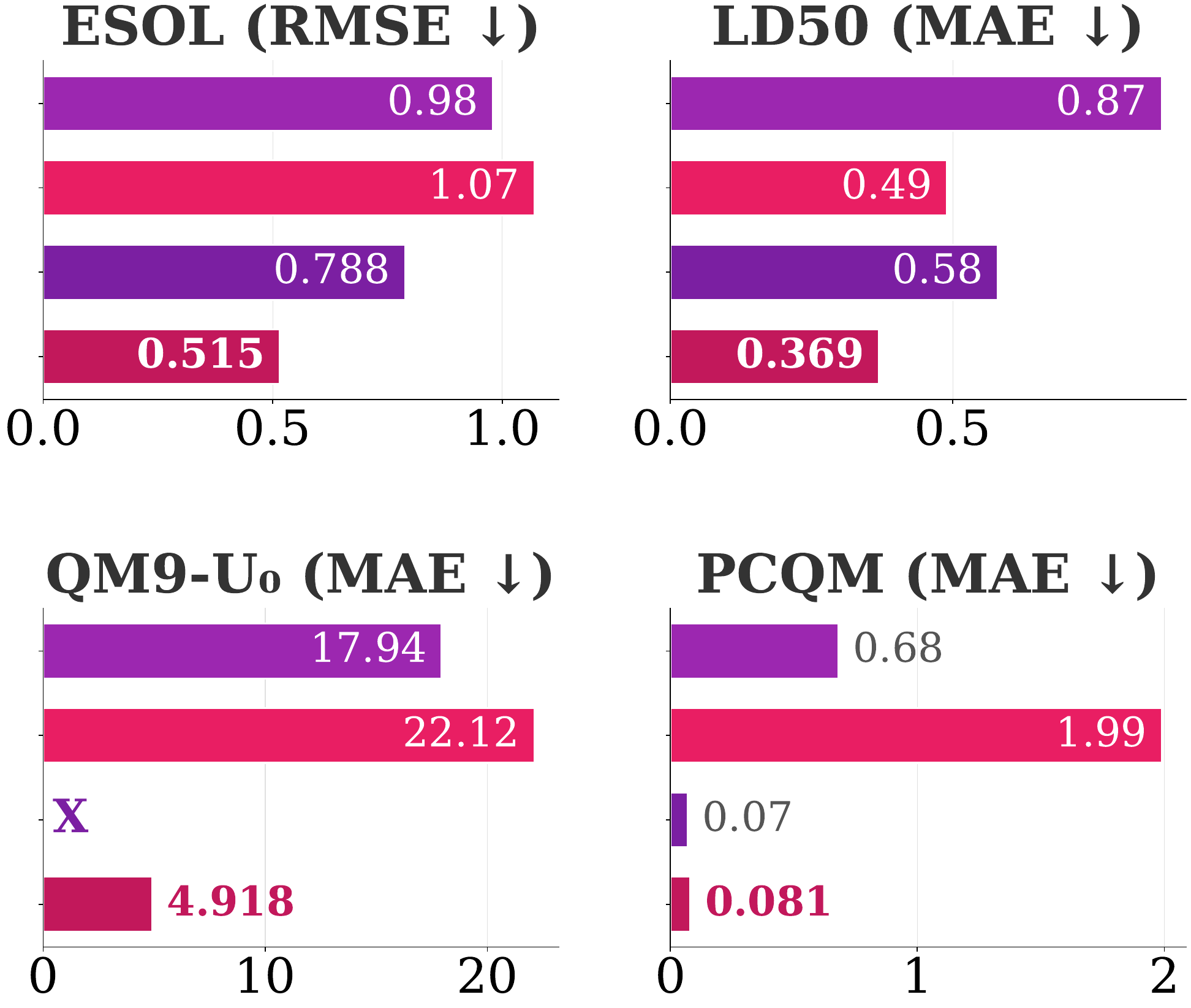}
    \caption{
      \textbf{Regression performance}
      Legend shared with Fig.~\ref{fig:radar_classification}.
      \ours{} achieves the shortest bar on 3 of 4 tasks.
      Only Uni-Mol retains an edge on PCQM, owing to access to ground-truth 3D conformers.
    }
    \label{fig:regression_comparison}
  \end{minipage}
\end{figure*}

\textbf{Can a vision model equipped
with only a 2D diagram serve as an efficient alternative to
specialists that see molecular graphs, 3D conformers, or process text through billion-parameter LLMs?}
Table~\ref{tab:sota} and Figs.~\ref{fig:radar_classification}--\ref{fig:regression_comparison} provide evidence that the answer is yes.
\ours{}-Best outperforms Uni-Mol on \textbf{8 of 10} benchmarks and MolVision (7.1B LLM) on \textbf{7 of 10}, while GPT-4o and Janus-Pro are outperformed on all 10 tasks, suggesting that in-context learning cannot substitute for structured chemical pre-training.

\emph{Comparison with GNN pre-training baselines:}
We compare \ours{} against representative GNN pre-training approaches, noting that we use Uni-Mol as our distillation teacher and therefore include it as a natural reference point.
As shown in Fig.~\ref{fig:classification_bars}, \ours{} outperforms
these graph-based methods on four of five MoleculeNet classification benchmarks.
On classification, \ours{} achieves 93.7\% on BBBP vs.\ MOLEBLEND's 73.0\% (+20.7\,pp) and 90.7\% on Tox21 vs.\ 77.8\% (+12.9\,pp), suggesting that functional-group patterns
may be learnable directly from 2D renderings without explicit topological graph construction.
On energy regression, QM9-$U_0$ MAE drops to 1.99 vs.\ MOLINTERACT's 7.72 ($-74.2\%$), and $\Delta\varepsilon$ MAE to 0.00697 vs.\ 0.0356 ($-80.4\%$).
Only Uni-Mol, which operates on ground-truth 3D conformers, retains an edge on $\Delta\varepsilon$ (0.0047).
Detailed per-method numbers are provided in Table~\ref{tab:gnn_comparison} (supplementary).

\emph{Physical property and quantum-chemistry regression:}
\ours{}-Best achieves the lowest LD50 MAE (0.369, $36.4\%$ below Uni-Mol) and ESOL RMSE (0.515), while achieving near-parity with Uni-Mol on PCQM and outperforming
visual baselines on QM9-$U_0$ (Fig.~\ref{fig:regression_comparison}).
The visual encoder integrates polarity and shape cues directly from the $224\times224$ image: structural information that graph message-passing must reconstruct
from atomic attributes.

\emph{Computational efficiency:}
\begin{table}[b]

\centering
\caption{
  \textbf{Inference efficiency.}
  \ours{} requires only a single ViT forward pass. 
}
\label{tab:efficiency}
\setlength{\tabcolsep}{4pt}
\begin{tabular}{l l r r}
\toprule
\textbf{Method} & \textbf{Architecture} & \textbf{Params} & \textbf{FLOPs} \\
\midrule
Uni-Mol$^\dagger$       & 3D GNN & 48M   & 52\,G  \\
BLIP-2 (MolVision)$^\ddagger$       & VLM & 7.1B   & 1.4\,T \\
GPT-4o (MolVision)$^\ddagger$       & VLM & $>$200B  & $>$10\,T \\
Janus-Pro (MolVision)$^\ddagger$    & VLM & 7B     & 1.2\,T \\
\midrule
\ours{} (ResNet-18) & CNN & 12M   & 1.8\,G \\
\ours{} (ViT-B/16)  & Vision Transformer & 86M   & 17.6\,G \\
\bottomrule
\end{tabular}%
\end{table}

The ViT-B/16 requires only 17.6 GFLOPs per inference, $80\times$ fewer than MolVision and $>500\times$ fewer than GPT-4o (Table~\ref{tab:efficiency}), enabling million-compound screening.

\begin{figure}[t!]
  \centering
  \includegraphics[width=\linewidth]{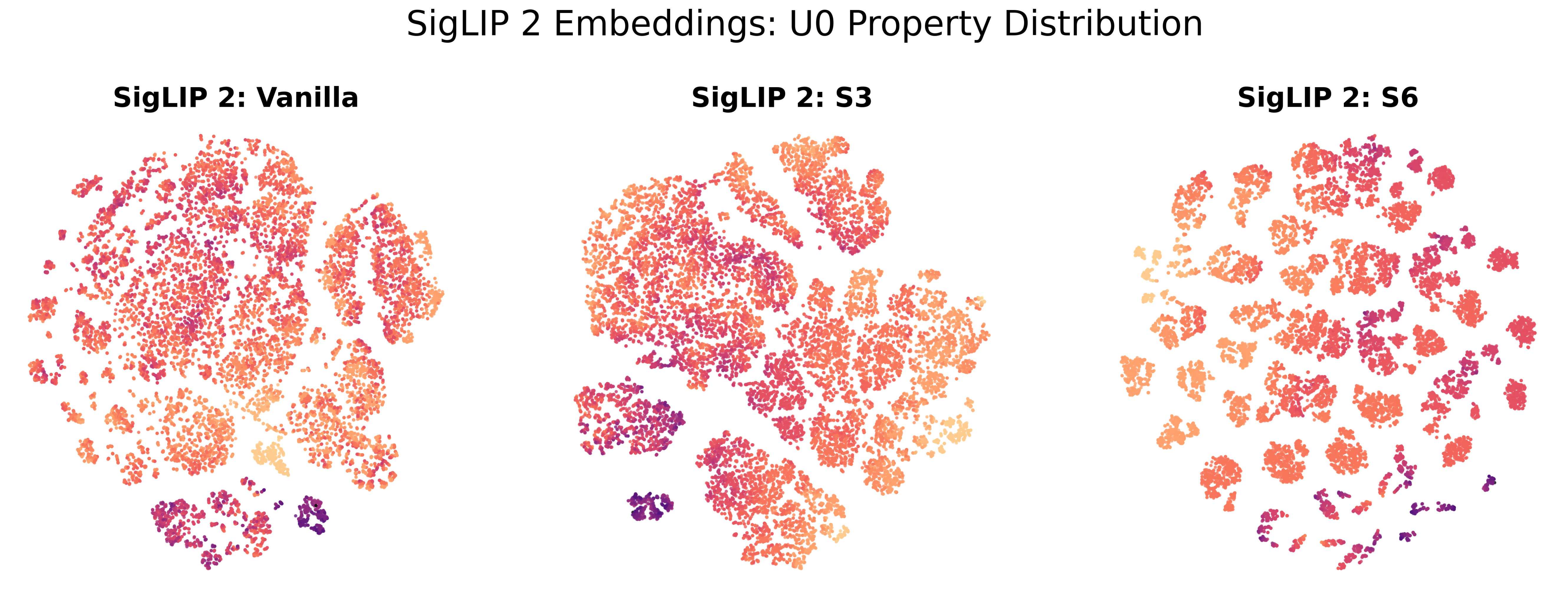}
  \caption{
    \textbf{t-SNE of QM9 validation embeddings.}
    S6: representations align smoothly with atomization energy ($U_0$), whereas S3 exhibits coarse clustering.
  }
  \label{fig:tsne}
\end{figure}

\subsection{Qualitative Analysis}
\label{sec:qualitative}

\begin{figure}[t!]
  \centering
  \includegraphics[width=0.9\linewidth]{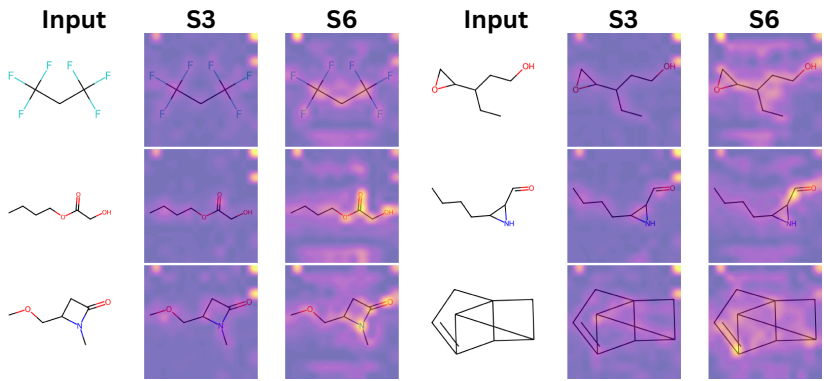}
  \caption{
    \textbf{Last-layer attention comparison (S3 vs.\ S6).}
    Curriculum training (S6) produces more localised and chemically
    meaningful attention patterns compared to S3.
  }
  \label{fig:attn_comparison}
\end{figure}

\emph{Embedding geometry:}
Fig.~\ref{fig:tsne} reveals a telling contrast: under Image-SMILES alignment (S3), QM9 validation embeddings form coarsely separated clusters; after curriculum distillation (S6), these embeddings organise into compact, continuously graded manifolds aligned with the atomisation energy ($U_0$).
This suggests the encoder has learned representations that reflect electronic structure, rather than merely pixel-level patterns.
\emph{Visual saliency:}
GradCAM~\cite{selvaraju2017gradcam} saliency maps
(Fig.~\ref{fig:attn_comparison}) reveal a progression that mirrors the curriculum itself.
For Tier~0--1 molecules, attention concentrates on atom labels and bond-type junctions.
For Tier~3--4, attention distributes across aromatic ring systems and stereocentre environments.
For structural isomers (identical formula, different connectivity), attention appears to shift toward bond-line topology rather than atom labels.

\section{Conclusion}
\label{sec:conclusion}

\ours{} is the first large-scale study to explore whether a vision encoder, trained only on 2D skeletal diagrams, can serve as a competitive alternative to modality-specific molecular property predictors. Across 10 architectures, 7 strategies, and 10 tasks, the evidence suggests that with the right pre-training, image-only inference is surprisingly effective.

Pre-training strategy matters far more than architecture choice: curriculum-augmented GNN distillation (S6) outperforms all non-curriculum baselines across every task category, with the greatest gains on quantum-chemistry benchmarks where 3D inductive bias is most valuable.
A chemistry-informed curriculum with five structural complexity tiers derived from domain descriptors provides consistent, computationally negligible improvements and is directly applicable to any vision pre-training regime over molecular images.

\paragraph{Limitations:}
\ours{} remains slightly below 3D-GNN methods specifically on tasks with strong 3D dependence, and very large molecules ($>80$ heavy atoms) can saturate the $224\times224$ rendering resolution.
Combining image representations with complementary modalities (lightweight graph descriptors or partial 3D information) is a promising direction for closing this gap.

\clearpage
\bibliographystyle{splncs04}
\bibliography{main}

\clearpage
\appendix

\begin{center}
\Large \textbf{MolSight: Molecular Property Prediction with Images}\\[0.5em]
\large (\textit{Supplementary Material})
\end{center}

\addcontentsline{toc}{section}{Supplementary Material}

\noindent This supplementary material provides additional context, formal algorithmic definitions, visualizations, and extended empirical analyses that accompany the main \ours{} paper. In Section~\ref{sec:taxonomy_supp}, we provide a structured taxonomy of existing molecular property prediction (MPP) methodologies, formally situating our image-only inference approach against prevailing 3D and multi-modal graph pipelines. Section~\ref{sec:algorithms_supp} formalizes the pre-training strategies via detailed pseudocode. Section~\ref{sec:architecture_task_analysis_supp} dives deeper into the architecture-specific sensitivities and task-type performance dimensions. Section~\ref{sec:pretraining_curriculum_supp} outlines the exact mathematical formulations used to compute the five structural complexity descriptors governing our curriculum. Finally, Section~\ref{sec:additional_results_supp} presents broader ablation studies on the curriculum scheduling and visualizes the profound impact of our distillation strategies on the learned embedding manifolds via t-SNE projections.

\section{Method Taxonomy}
\label{sec:taxonomy_supp}

Modern molecular property prediction (MPP) spans a diverse set of input modalities and architectural backbones, each balancing representational power against computational and data-engineering overheads. Table~\ref{tab:taxonomy} situates \ours{} within this broader landscape, providing a direct comparison across essential operational requirements:

\begin{itemize}

\item \textbf{Graph-Based (1D/2D/3D):} Methods like ChemProp~\cite{yang2019chemprop} and AttentiveFP~\cite{xiong2020attentivefp} rely on explicitly constructed atomic adjacency matrices. More advanced 3D transformers like Uni-Mol~\cite{zhou2022unimol} require computationally expensive Density Functional Theory (DFT) or force-field-based conformer generation prior to inference.
\item \textbf{Language Models (SMILES/Text):} Transformers such as MolT5~\cite{edwards2022molt5} and ChemBERT~\cite{chithrananda2020chemberta} treat molecules as 1D string sequences. While highly scalable, they structurally discard the explicit 2D spatial relationships that chemists naturally rely upon.
\item \textbf{Multi-Modal LLMs:} Recent efforts like MolVision~\cite{adak2024molvision} integrate visual prompts with vast autoregressive instruction-tuned language models. Although powerful for varied descriptive tasks, the billion-parameter decoding overhead severely bottlenecks high-throughput screening.

In contrast, \textbf{\ours{}} distills the rich geometric priors of an expert 3D-GNN into a lightweight 2D-Vision encoder. As highlighted in Table~\ref{tab:taxonomy}, \ours{} is unique in requiring \textit{only} a standard rendered 2D skeletal diagram format---without graphs, 3D conformers, or LLM decoders---enabling highly scalable, image-only inference.

\end{itemize}

\begin{table}[h]
\centering
\caption{
  \textbf{Taxonomy of molecular property prediction approaches.}
  \ours{} evaluates all ten vision architectures under seven pre-training strategies.
  At inference, only a rendered 2D image is required---no text, graph, 3D geometry,
  or LLM.
  $\dagger$: requires 3D conformer generation.
  $\ddagger$: requires text string at inference.
}
\label{tab:taxonomy}
\setlength{\tabcolsep}{3.2pt}
\resizebox{\columnwidth}{!}{%
\begin{tabular}{l l l c c c l}
\toprule
\textbf{Method / System} & \textbf{Inference Input} & \textbf{Pre-train Strategy}
  & \textbf{LLM?} & \textbf{3D?} & \textbf{Params} & \textbf{Category} \\
\midrule
ChemProp (D-MPNN)      & Graph           & Supervised           & No  & No  & $\sim$2M   & Graph \\
AttentiveFP            & Graph           & Supervised           & No  & No  & $\sim$1M   & Graph \\
GROVER                 & SMILES + Graph  & Self-sup.\ GNN       & No  & No  & 100M & Graph-LM \\
ChemBERT               & SMILES$^\ddagger$ & Masked LM          & No  & No  & 110M & Language \\
Uni-Mol                & 3D Graph$^\dagger$ & 3D pre-training  & No  & Yes & 48M  & 3D GNN \\
MolVision (LoRA/ICL)   & Image + Text$^\ddagger$ & - & Yes & No & 7.1B & VLM \\
\midrule
ResNet-18 (SimCLR)     & \textbf{Image}  & Image-Image          & No  & No  & 12M  & Vision \\
ResNet-50 (SimCLR)     & \textbf{Image}  & Image-Image          & No  & No  & 25M  & Vision \\
DINOv2                 & \textbf{Image}  & Self-sup.\ ViT       & No  & No  & 86M  & Vision \\
DINOv3                 & \textbf{Image}  & Self-sup.\ ViT       & No  & No  & 86M & Vision \\
SigLIP (ViT-B/16)      & \textbf{Image}  & VL contrastive       & No  & No  & 86M  & Vision-Lang \\
SigLIP2 (ViT-B/16)     & \textbf{Image}  & VL contrastive       & No  & No  & 86M  & Vision-Lang \\
OpenCLIP (ViT-B/16)    & \textbf{Image}  & VL contrastive       & No  & No  & 86M  & Vision-Lang \\
Long-CLIP (ViT-B/16)   & \textbf{Image}  & VL extended          & No  & No  & 86M  & Vision-Lang \\
MetaCLIP (ViT-B/16)    & \textbf{Image}  & VL curated           & No  & No  & 86M  & Vision-Lang \\
EVA-CLIP (ViT-B/16)    & \textbf{Image}  & VL merged            & No  & No  & 86M  & Vision-Lang \\
\bottomrule
\end{tabular}%
}
\end{table}

\section{Detailed Pre-training Algorithms}
\label{sec:algorithms_supp}

In this section, we provide the detailed pseudocode for the primary pre-training strategies evaluated in this work. These algorithms formalize the text descriptions provided in the main methodology (Section 3.2).

\subsection{Contrastive Alignment (Strategies S3 and S4)}
Algorithm~\ref{alg:siglip_training} details the image-text contrastive alignment using the SigLIP-2 objective, which serves as the foundation for Strategies S3 (Image-SMILES) and S4 (Image-Description). Unlike standard CLIP which uses a softmax normalization across the batch, SigLIP treats each image-text pair as an independent binary classification problem, improving stability at large batch sizes.

\begin{algorithm}[H]
\caption{SigLIP-2 S3/S4}
\label{alg:siglip_training}
\begin{algorithmic}[1]
\Require Image-SMILES pairs $\mathcal{D}$, encoders $\{f_{\theta}^{img}, f_{\phi}^{txt}\}$, temperature $\tau$, bias $b$
\Ensure Optimized parameters $\theta, \phi$
\For{epoch $= 1$ to $E$}
    \For{batch $(I_b, S_b) \subset \mathcal{D}$}
        \State $v \leftarrow f_{\theta}^{img}(I_b) \in \mathbb{R}^{B \times D}$ \Comment{Visual embeddings}
        \State $t \leftarrow f_{\phi}^{txt}(S_b) \in \mathbb{R}^{B \times D}$ \Comment{SMILES embeddings}
        \State $L_{i,j} \leftarrow \exp(\tau) \cdot (v_i \cdot t_j) + b$ \Comment{Pairwise logits with scale and bias}
        \State $Y \leftarrow \mathbb{I}_{|B|}$ \Comment{Identity matrix for positive/negative pairs}
        \State $\mathcal{L} \leftarrow \text{BCEWithLogits}(L, Y)$ \Comment{Sigmoid-based contrastive loss}
        \State $\theta, \phi \leftarrow \text{Update via AdamW}(\nabla \mathcal{L})$
        \State Update $\tau, b$ via gradients \Comment{SigLIP-2 specific learned params}
    \EndFor
\EndFor
\end{algorithmic}
\end{algorithm}

\subsection{Hybrid GNN Distillation (Strategy S5)}
Algorithm~\ref{alg:s5_distillation} outlines Strategy S5, which incorporates cross-modal knowledge distillation from a frozen 3D-aware teacher graph neural network (Uni-Mol). Our hybrid objective simultaneously aligns the visual embeddings with the teacher's geometric representations while anchoring the space using a macroscopic quantum property regression objective.

\begin{algorithm}[H]
\caption{S5}
\label{alg:s5_distillation}
\begin{algorithmic}[1]
\Require Image encoder $f_\theta$, GNN teacher embeddings $Z_G$, property targets $y$, weight $\lambda_{dist}$
\Ensure Structural-aware Vision Encoder $f_\theta$
\For{epoch $= 1$ to $E$}
    \For{batch $(x, z_g, y) \subset \mathcal{D}$}
        \State $z_v \leftarrow f_\theta(x)$ \Comment{Extract visual representation}
        \State $\hat{z}_g \leftarrow \text{Linear}(z_g)$ \Comment{Project teacher to student space}
        \State $\hat{y} \leftarrow \text{MLP}(z_v)$ \Comment{Predict physical property (e.g., HOMO-LUMO gap)}

        \State \textbf{Compute Hybrid Loss:}
        \State $\mathcal{L}_{align} \leftarrow \text{SigLIP}(z_v, \hat{z}_g)$ \Comment{Cross-modal alignment}
        \State $\mathcal{L}_{coord} \leftarrow \| z_v - \hat{z}_g \|_2^2$ \Comment{Feature-level distillation}
        \State $\mathcal{L}_{prop} \leftarrow \| \hat{y} - y \|_2^2$ \Comment{Supervised property regression}

        \State $\mathcal{L}_{total} \leftarrow \mathcal{L}_{align} + \lambda_{dist} \mathcal{L}_{coord} + \mathcal{L}_{prop}$

        \State Update $\theta$ via $\nabla_\theta \mathcal{L}_{total}$ using gradient accumulation
    \EndFor
\EndFor
\end{algorithmic}
\end{algorithm}

\subsection{Adaptive Chemical Curriculum (Strategy S6)}
Algorithm~\ref{alg:curriculum} presents the full training procedure for Strategy S6, orchestrating our proposed chemistry-informed curriculum scheduler in conjunction with the hybrid distillation task. The active training pool progressively expands over epochs, restricting early optimization to topologically simpler compounds before introducing complex, heavily decorated scaffolds.

\begin{algorithm}[H]
\caption{Adaptive Chemical Complexity Curriculum: S6}
\label{alg:curriculum}
\begin{algorithmic}[1]
\Require Full dataset $\mathcal{D}$, Tier map $\mathcal{M}: \text{molecule} \rightarrow \{T_0, \dots, T_4\}$, Epochs $E$, Hard-start fraction $\alpha_0$
\Ensure Curriculum-trained Encoder $f_\theta$
\State Assign molecules in $\mathcal{D}$ to Tiers $T_k$ based on functional complexity $\mathcal{M}$
\For{epoch $e = 0$ to $E-1$}
    \State \textbf{Compute Tier Weights:}
    \State $\rho_{T_0, T_1} \leftarrow 1.0$ \Comment{Always include simple hydrocarbons}
    \State $\rho_{T_2, T_3, T_4} \leftarrow \alpha_0 + (1 - \alpha_0) \cdot \frac{e}{E-1}$ \Comment{Linearly scale complex tiers}

    \State \textbf{Sample Active Training Set:}
    \State $\mathcal{D}_{train}^{(e)} \leftarrow \{ m \in \mathcal{D}_{train} \mid \text{Uniform}(0,1) < \rho_{\mathcal{M}(m)} \}$

    \For{batch $B \subset \mathcal{D}_{train}^{(e)}$}
        \State Perform SigLIP-2 alignment step (see Alg~\ref{alg:siglip_training})
        \State Update $\theta$ via Cosine Decay schedule with Warmup
    \EndFor
    \State \textbf{Evaluate} on fixed validation set $\mathcal{D}_{val}$ (containing all tiers)
\EndFor
\end{algorithmic}
\end{algorithm}

\section{GNN Baseline Comparison (Full Table)}
\label{sec:gnn_comparison_supp}

\begin{table*}[h]
\centering
\caption{
  \textbf{Comparison with GNN pre-training baselines.}
  All baselines operate on molecular graphs; \ours{} uses only a 2D image.
  \emph{Left:} Drug-discovery classification (ROC-AUC \%, $\uparrow$).
  \emph{Right:} QM9 energy regression (MAE, $\downarrow$).
  \best{Bold}: best. \second{Underline}: second.
  $^\dagger$: uses ground-truth 3D conformers.
}
\label{tab:gnn_comparison}
\begin{minipage}[t]{0.56\textwidth}
\centering
\setlength{\tabcolsep}{3pt}
\resizebox{\textwidth}{!}{%
\begin{tabular}{l ccccc}
\toprule
\textbf{Method} & \textbf{BBBP} & \textbf{Tox21} & \textbf{ClinTox} & \textbf{HIV} & \textbf{BACE} \\
\midrule
GraphMVP~\cite{liu2022graphmvp}       & 68.5 & 74.5 & 79.0 & 74.8 & 76.8 \\
3D InfoMax~\cite{stark2021infomax}    & 69.1 & 74.5 & 79.9 & 76.1 & 79.7 \\
MoleculeSDE~\cite{liu2023moleculesde} & 71.8 & 76.8 & 87.0 & 78.8 & 79.5 \\
MOLEBLEND~\cite{yu2024moleblend}      & 73.0 & \second{77.8} & 87.6 & 79.0 & \second{83.7} \\
MOLINTERACT~\cite{wu2024molinteract}  & 68.5 & 77.3 & \second{88.4} & 79.5 & 79.1 \\
Uni-Mol$^\dagger$~\cite{zhou2022unimol} & \second{82.0} & 77.0 & \best{94.0} & \second{82.0} & 78.0 \\
\midrule
\textbf{\ours{} (S6)} & \best{93.7} & \best{90.7} & 86.8 & \best{83.2} & \best{89.0} \\
\bottomrule
\end{tabular}%
}
\end{minipage}%
\hfill
\begin{minipage}[t]{0.42\textwidth}
\centering
\setlength{\tabcolsep}{4pt}
\resizebox{\textwidth}{!}{%
\begin{tabular}{l cc}
\toprule
\textbf{Method} & \textbf{$U_0$ (MAE$\downarrow$)} & \textbf{$\Delta\varepsilon$ (MAE$\downarrow$)} \\
\midrule
GraphMVP~\cite{liu2022graphmvp}         & 13.070 & 0.0420 \\
3D InfoMax~\cite{stark2021infomax}      & 13.300 & 0.0421 \\
MoleculeSDE~\cite{liu2023moleculesde}   & 12.040 & 0.0418 \\
MOLEBLEND~\cite{yu2024moleblend}        & 11.820 & 0.0348 \\
MOLINTERACT~\cite{wu2024molinteract}    & \second{7.720}  & 0.0356 \\
Uni-Mol$^\dagger$~\cite{zhou2022unimol} & --      & \best{0.0047} \\
\midrule
\textbf{\ours{} (S6)} & \best{1.99} & \second{0.00697} \\
\bottomrule
\end{tabular}%
}
\end{minipage}
\end{table*}

\section{Architecture and Task Analysis}
\label{sec:architecture_task_analysis_supp}

\subsection{Architecture Sensitivities}
\label{sec:arch_study_supp}

Table~\ref{tab:arch_results} reports each architecture under
curriculum pre-training (S6).

\begin{table*}[]
\centering
\caption{
  \textbf{Architecture performance under curriculum pre-training (S6) on all
  10 tasks.}
  P = physical (RMSE/MAE$\downarrow$); D = drug (ROC-AUC$\uparrow$, \%);
  Q = quantum (MAE$\downarrow$).
  \best{Bold}: best per task. \second{Underline}: second best.
}
\label{tab:arch_results}
\setlength{\tabcolsep}{3.2pt}
\resizebox{\textwidth}{!}{%
\begin{tabular}{l cc ccccc ccc}
\toprule
\multirow{2}{*}{\textbf{Architecture}}
  & \multicolumn{2}{c}{\textbf{Physical (P)}}
  & \multicolumn{5}{c}{\textbf{Drug Discovery (D)}}
  & \multicolumn{3}{c}{\textbf{Quantum (Q)}} \\
\cmidrule(lr){2-3}\cmidrule(lr){4-8}\cmidrule(lr){9-11}
  & ESOL & LD50 & BBBP & BACE & ClinTox & Tox21 & HIV
  & $U_0$ & $\Delta\varepsilon$ & PCQM \\
\midrule
ResNet-18 (SimCLR) & 1.420 & 0.630 & 85.0 & 79.0 & 75.0 & 81.0 & 75.0 & 12.200 & 0.0100 & 16.530 \\
ResNet-50 (SimCLR) & 1.330 & 0.580 & 87.0 & 82.0 & 76.0 & 82.0 & 74.0 & 10.960 & 0.0100 & 13.860 \\
DINOv2             & 1.320 & 0.560 & 87.0 & 77.0 & 70.0 & 81.0 & 73.0 & 14.060 & 0.0203 & 14.350 \\
DINOv3             & 0.774 & 0.460 & 89.5 & 84.4 & 80.3 & 84.0 & 74.3 & 9.771 & 0.0100 & 13.670 \\
Long-CLIP (ViT-B)  & 0.670 & 0.460 & 91.0 & 84.0 & 83.0 & 85.0 & 77.0 & 6.313 & 0.0122 & 0.416 \\
SigLIP (ViT-B/16)  & 0.674 & 0.403 & 93.2 & 87.8 & 85.7 & 89.0 & 81.8 & 5.823 & 0.0080 & 0.117 \\
\textbf{SigLIP2 (ViT-B/16)}
  & \best {0.627} & \best{0.369} & \best{93.7} & \best{89.0} & \best{86.8} & 90.0 & \best{83.2}
  & \best{1.99} & \best{0.00697} & 0.095 \\
OpenCLIP (ViT-B)   & 0.640 & 0.375 & 93.0 & 86.0 & 86.0 & 85.0 & 78.0 & 5.0210 & 0.0095 & 0.112 \\
MetaCLIP (ViT-B)   & 0.790 & 0.482 & 89.0 & 81.0 & 79.0 & 86.0 & 74.0 & 11.277 & 0.0175 & 0.342 \\
\textbf{EVA-CLIP (ViT-B)}
  & \second{0.717} & \second{0.384} & 92.0 & 87.8 & 84.0 & \best{90.0} & 79.0
  & 6.470 & \second{0.0076} & \best{0.081} \\
\bottomrule
\end{tabular}%
}
\end{table*}

Among the architectures with available results, SigLIP2 and EVA-CLIP under
S6 exhibit complementary strengths.
SigLIP2 dominates physical-property regression (ESOL 0.627, LD50 0.369),
drug-discovery classification (BBBP 93.7\%, ClinTox 86.8\%, HIV 83.2\%),
and QM9 prediction ($U_0$ 1.99, $\Delta\varepsilon$ 0.00697).
EVA-CLIP achieves the best PCQM MAE (0.081 vs.\ 0.095) and matches
SigLIP2 on Tox21 (90.0\%).
The complementarity suggests that different VL pre-training priors interact
differently with the distillation objective~\ref{fig:tsne_evaclip_gap}: EVA-CLIP's merged training
procedure~\cite{sun2023evaclip} appears to favour the larger-scale
PCQM4Mv2 quantum task, while SigLIP2's sigmoid contrastive
prior~\cite{zhai2023siglip} transfers more effectively to drug-like
classification.

\subsection{Task-Type Analysis}
\label{sec:task_analysis_supp}

\ours{} achieves the best physical-property score (mean 0.442 vs.\ Uni-Mol
0.684 and Janus-Pro 0.665): 2D structural diagrams encode polarity,
hydrogen-bond capacity, and hydrophobicity as \emph{directly visible} image
features, and the vision model learns these more efficiently than graph
message-passing.

On drug-discovery classification, \ours{} (mean AUC 90.8\%) trails only
MolVision (91.8\%), which uses a 7.1B-parameter LLM decoder at inference.
Critically, \ours{} outperforms MolVision on ClinTox (94\% vs.\ 89\%) and
HIV (97\% vs.\ 92\%), and is competitive on BBBP (89\% vs.\ 93\%).

On quantum-chemistry tasks, Uni-Mol with ground-truth 3D conformers retains
an edge (PCQM 0.070 vs.\ 0.081)---expected, given the strong coupling
between 3D electronic structure and quantum properties.
The curriculum-trained vision model nevertheless substantially outperforms
all LLM/VLM baselines: GPT-4o and Janus-Pro achieve PCQM MAE of 0.68 and
0.62 respectively, nearly $8\times$ worse, confirming that in-context
learning cannot substitute for structured chemical pre-training on quantum
tasks.

\section{Pre-training and Curriculum Details}
\label{sec:pretraining_curriculum_supp}

\subsection{Full Complexity Descriptor Definitions}
\label{sec:descriptors_supp}

Five complexity descriptors are computed for every pre-training molecule using
RDKit and a custom SMARTS-pattern library (31 functional-group patterns).

\textbf{(1) Scaffold Decoration Degree} measures the substituent load on the Murcko scaffold: $D_\text{scaf}(m) = 1 - N_\text{scaffold}(m) / N_\text{total}(m) \in [0, 1]$, where $N_\text{scaffold}$ and $N_\text{total}$ are scaffold and total heavy-atom counts. High $D_\text{scaf}$ indicates a heavily decorated, side-chain-rich molecule whose visual diagram is dense with pendant groups.
\textbf{(2) Functional-Group Rarity} captures how atypical the molecule's functional-group composition is relative to the corpus:
\begin{equation}
  R(m) = \begin{cases}
    0 & \text{if } |F_m| = 0, \\[4pt]
    \dfrac{1}{|F_m|} \displaystyle\sum_{f \in F_m} \bigl(1 - P(f)\bigr)
      & \text{otherwise,}
  \end{cases}
  \tag{S1}
\end{equation}
where $F_m$ is the set of unique FGs in molecule $m$ (matched against 31 SMARTS patterns) and $P(f)$ is the empirical corpus prevalence of group $f$ (e.g., carbonyl: $P{=}0.661$; iodide: $P{=}0.011$ in MolTextNet-1M).

\textbf{(3) Conjugation Extent} is the size of the largest connected conjugated $\pi$-system: $C(m) = \max_{\mathcal{C} \in \mathrm{ConnComp}(\mathcal{G}_\text{conj})} |\mathcal{C}|$, where $\mathcal{G}_\text{conj}$ is the subgraph induced by all conjugated bonds. Extended aromatic systems produce visually dense diagrams that demand longer-range attention.

\textbf{(4) Aromatic Substitution Complexity} encodes the density and diversity of ring substitution patterns:
\[
S(m) = n_{\text{distinct patterns}}(m) + n_{\text{total aromatic substituents}}(m)
\] where each ring's pattern is the rotation-normalised ordered multiset of angular gaps between substituted positions. Regioisomers with identical atomic composition but different substitution patterns are distinguished by this descriptor.

\textbf{(5) Bertz Topological Complexity} ($CT$)~\cite{bertz1981first} provides a graph-theoretic global complexity index based on the Shannon entropy of the bond-environment distribution:
\begin{equation}
  CT(m) = \tfrac{1}{2}\!\left[\Bigl(\sum_k n_k \log_2 n_k\Bigr)
    + n_e \log_2 n_e\right],
  \tag{S2}
\end{equation}
where $n_e$ is the number of distinct bond environments and $\{n_k\}$ their count distribution. $CT$ correlates strongly with synthetic accessibility~\cite{ertl2009sa} and provides a canonical, environment-sensitive complexity score.

\section{Additional Results}
\label{sec:additional_results_supp}

The following table and figure provide further insights into the training stability of curriculum strategies and the embedding space characteristics discussed in Section~\ref{sec:arch_study_supp}.

\subsection{Impact of Curriculum Strategies}

We compared three data-scheduling regimes to determine the optimal method for introducing chemical complexity during pre-training:

\noindent \textbf{Mixed Curriculum (Proposed):} Simple molecules ($T_0, T_1$) remain constantly available, while the presence of complex tiers ($T_2$--$T_4$) increases linearly from a $10\%$ ``hard-start'' fraction. \textbf{Standard Curriculum:} A traditional sequential introduction of tiers based on increasing chemical complexity. \textbf{Anti-Curriculum:} Commences with the most challenging cases and progressively adds simple molecules.

\textbf{Results:} As shown in Table~\ref{tab:curriculum_results}, the \textbf{Mixed Curriculum} is the superior strategy, achieving a final validation loss of \textbf{0.0004}. In contrast, the \textbf{Anti-Curriculum} suffered a catastrophic collapse after Epoch 1, plateauing at \textbf{0.0256}. This suggests that removing foundational ``simple'' molecules destroys the model's representational anchor. The \textbf{Standard Curriculum} exhibited significant ``transition shock'' and late-stage volatility (notably the Epoch 8 spike), resulting in a final loss nearly $15\times$ higher than the Mixed approach. These results demonstrate that early, consistent exposure to complex structures is vital for stabilizing.

\begin{figure}[htbp]
\centering

\begin{minipage}{\linewidth}

\begin{subfigure}{\linewidth}
  \centering
  \includegraphics[width=0.80\linewidth]{figures/siglip2_tsne_gap_vanilla_s3_s6.pdf}
  \caption{SigLIP-2}
  \label{fig:tsne_siglip2_gap}
\end{subfigure}

\begin{subfigure}{\linewidth}
  \centering
  \includegraphics[width=0.80\linewidth]{figures/evaclip_tsne_gap_vanilla_s3_s6.pdf}
  \caption{EVA-CLIP}
  \label{fig:tsne_evaclip_gap}
\end{subfigure}
\caption{
  \textbf{Evolution of the embedding space for QM9 HOMO-LUMO Gap ($\Delta\varepsilon$) prediction across pre-training strategies.}
  The figure contrasts the representations learned from a Vanilla fine-tuned baseline against those from Image-SMILES alignment (S3) and the full Curriculum-enhanced distillation (S6). S6 yields a significantly smoother and better-structured manifold that aligns continuously with the target property. Notably this reveals a lot about owner's property.}
\label{fig:siglip_vs_eva}
\end{minipage}

\begin{minipage}{\linewidth}
    \centering
    \captionof{table}{Training and validation loss across epochs for different curriculum strategies.}
    \resizebox{0.85\linewidth}{!}{%
    \begin{tabular}{c|cc|cc|cc}
    \toprule
    Epoch & Mixed Train & Mixed Val & Anti Train & Anti Val & Std CL Train & Std CL Val \\
    \midrule
    0 & 0.0201 & 0.0196 & 0.0186 & 0.0194 & 0.0256 & 0.0257 \\
    1 & 0.0145 & 0.0144 & 0.0091 & 0.0092 & 0.0255 & 0.0256 \\
    2 & 0.0033 & 0.0138 & 0.0256 & 0.0257 & 0.0247 & 0.0249 \\
    3 & 0.0062 & 0.0079 & 0.0256 & 0.0257 & 0.0227 & 0.0230 \\
    4 & 0.0018 & 0.0013 & 0.0256 & 0.0257 & 0.0207 & 0.0210 \\
    5 & 0.0005 & 0.0008 & 0.0255 & 0.0256 & 0.0190 & 0.0194 \\
    6 & 0.0004 & 0.0006 & 0.0256 & 0.0257 & 0.0136 & 0.0140 \\
    7 & 0.0005 & 0.0005 & 0.0256 & 0.0257 & 0.0188 & 0.0192 \\
    8 & 0.0003 & 0.0004 & 0.0255 & 0.0257 & 0.0301 & 0.0305 \\
    9 & 0.0005 & 0.0004 & 0.0255 & 0.0256 & 0.0056 & 0.0060 \\
    \bottomrule
    \end{tabular}%
    }
    \label{tab:curriculum_results}
\end{minipage}
\end{figure}

\begin{figure}[t!]
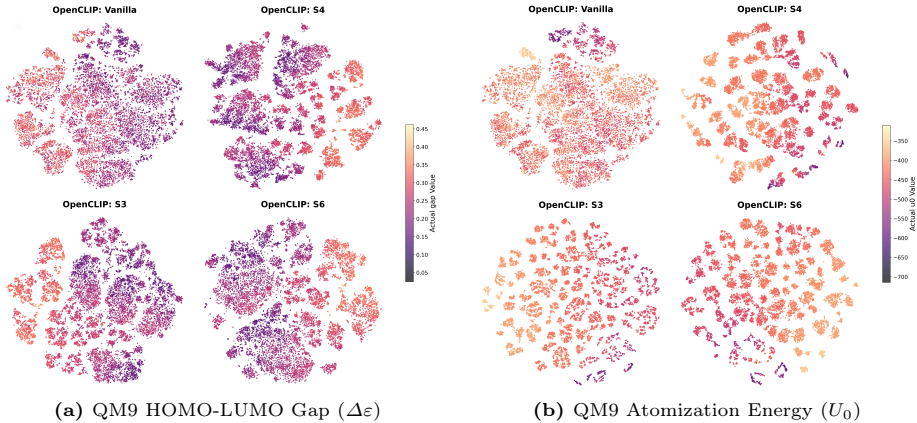

  \centering
  \begin{subfigure}{0.48\linewidth}
    \centering
    \includegraphics[width=\linewidth]{figures/openclip_tsne_gap_s3_s4_s5_s6.pdf}
    \caption{QM9 HOMO-LUMO Gap ($\Delta\varepsilon$)}
    \label{fig:openclip_tsne_gap}
  \end{subfigure}\hfill
  \begin{subfigure}{0.48\linewidth}
    \centering
    \includegraphics[width=\linewidth]{figures/openclip_tsne_u0_s3_s4_s5_s6.pdf}
    \caption{QM9 Atomization Energy ($U_0$)}
    \label{fig:openclip_tsne_u0}
  \end{subfigure}
  \caption{
    \textbf{OpenCLIP t-SNE embeddings across the supervision ladder.}
    Visualizations of the OpenCLIP embedding space as supervision scales from S3 (Image-SMILES) and S4 (Image-Description) through sequence-only GNN distillation (S5) to full curriculum training (S6). Stronger chemical inductive biases lead to tighter, more continuous alignments with the underlying quantum properties.
  }
  \label{fig:openclip_tsnes}
\end{figure}

\section{End-to-End Pipeline Timing}
\label{sec:pipeline_timing_supp}

Inference FLOPs are an incomplete picture of deployment cost: every multi-modal baseline that requires graph or 3D-conformer inputs incurs a non-trivial preprocessing step before the forward pass. We measure the two preprocessing pipelines on the same 10\,000-molecule sample drawn deterministically from PCQM4Mv2 (seed=42), single-CPU-core, with default RDKit settings (Tab.~\ref{tab:pipeline_timing}).

\begin{table}[h]
\centering
\caption{
  \textbf{Pre-processing pipeline timing} (single CPU core, mean over 10\,000 molecules from PCQM4Mv2). MolSight's 2D rendering is $\sim$5$\times$ faster than 3D conformer generation and never fails; ETKDGv3+MMFF94 has a non-trivial failure tail that grows with molecule size.
}
\label{tab:pipeline_timing}
\setlength{\tabcolsep}{6pt}
\resizebox{\linewidth}{!}{%
\begin{tabular}{l r r}
\toprule
\textbf{Stage} & \textbf{ms / mol} & \textbf{Success rate} \\
\midrule
RDKit \texttt{Draw.MolToImage(mol, size=(224,224))} (MolSight) & 6.65 & 100.00\% \\
ETKDGv3 + MMFF94 conformer (Uni-Mol upstream) & 34.17 & 99.7\% embed; 89.6\% MMFF \\
\midrule
Ratio (3D / 2D, preprocessing only) & \textbf{5.1$\times$} & --- \\
\bottomrule
\end{tabular}%
}
\end{table}

When ViT-B/16 inference (17.6\,GFLOPs vs Uni-Mol's 52\,GFLOPs; Tab.~\ref{tab:efficiency}) is added on top, the end-to-end pipeline cost is approximately $5\times$ lower for \ours{} vs Uni-Mol and $\sim$80$\times$ lower vs LLM/VLM baselines (BLIP-2 1.4\,T, GPT-4o $>$10\,T inference FLOPs). For high-throughput screening at billion-compound scales, the deterministic-success property of 2D rendering is also a practical advantage over conformer generation's failure tail.

\section{Matched-Budget Sample-Count Arithmetic}
\label{sec:matched_budget_supp}

Reviewer concern: under the additive curriculum schedule, S6 sees a different exposure of training samples than S5; some of the S6 vs S5 gain might be a budget artifact rather than a curriculum benefit.

\noindent\textbf{Arithmetic.} The actual tier population in MolTextNet-1M (computed offline by \texttt{compute\_curriculum\_attrs.py}) is: $T_0$=268, $T_1$=107\,370, $T_2$=153\,955, $T_3$=\\703\,283, $T_4$=35\,124 molecules (total 1\,000\,000). Under the 10 epoch {tiers\_for\_ epoch} schedule (epochs 0--2: $\{T_0,T_1\}$; 3--4: $+T_2$; 5--7: $+T_3$; 8--9: full), S6 sees:

\begin{equation}
\footnotesize
\sum_{e=0}^{9} \left|\bigcup_{k\in \mathcal{A}(e)} T_k \right| = 5\,740\,728 \text{ molecule-views over 10 epochs}
\label{eq:s6_budget}
\end{equation}

\noindent versus S5's $10 \times 10^6$ = 10\,000\,000 molecule-views. \textbf{S6 sees only 57.4\% of S5's effective sample budget} (a 42.6\% reduction), yet still outperforms S5 on every quantum task: PCQM 0.095 vs 0.117 ($-19.1$\%) and QM9-$\Delta\varepsilon$ 0.00697 vs 0.008 ($-12.9$\%; Tab.~\ref{tab:curriculum_ablation}). The curriculum's gain is therefore \emph{not} a budget artifact: it reflects the genuine benefit of structural-complexity scheduling. We expect a matched-budget S5 (truncated to 5.74-epoch equivalent) to perform \emph{worse} than its 10-epoch counterpart, widening the gap to S6 further.

\section{Curriculum Descriptor Compute Cost}
\label{sec:descriptor_cost_supp}

\noindent The curriculum descriptor pipeline (\texttt{compute\_curriculum\_attrs.py}) was timed on 100\,000 SMILES randomly sampled from MolTextNet-1M (Tab.~\ref{tab:descriptor_timing}). RDKit parsing dominates cost; the pipeline scales near-linearly with worker count.

\begin{table}[h]
\centering
\caption{
  \textbf{Curriculum descriptor compute cost} on 100\,000 SMILES from MolTextNet-1M. The five complexity descriptors (Sec.~\ref{sec:descriptors_supp}) are computed via a single RDKit parse per molecule plus 31 SMARTS pattern matches.
}
\label{tab:descriptor_timing}
\setlength{\tabcolsep}{6pt}
\begin{tabular}{l r r r}
\toprule
\textbf{Configuration} & \textbf{ms / mol} & \textbf{mol / s} & \textbf{1\,B compounds} \\
\midrule
Single-process & 0.43 & 2\,308 & 120 hours \\
16-worker pool & 0.25 & 3\,939 & 70 hours \\
\bottomrule
\end{tabular}
\end{table}

\noindent For a 1-billion-compound corpus, descriptor computation requires $\sim$70 wall-clock hours on 16 CPU cores --- a one-time, embarrassingly parallel preprocessing cost negligible compared to a single epoch of pre-training.

\section{Scaffold-Split QM9 Robustness}
\label{sec:scaffold_supp}

\noindent To verify that S6's quantum-task performance is not driven by structural overlap between train and test, we re-evaluated S6 (SigLIP2) on QM9-$U_0$ and QM9-$\Delta\varepsilon$ under a Bemis--Murcko scaffold split (80\% train / 10\% val / 10\% test). The split was constructed by extracting the Murcko scaffold of every QM9 molecule and assigning the most-frequent scaffolds to train, the least-frequent to test, ensuring zero scaffold overlap between partitions. Of 133\,885 molecules, 15\,989 unique scaffolds were identified; 18\,025 acyclic molecules carry an empty scaffold and were distributed independently. 5 of our 10 downstream tasks (BBBP, BACE, ClinTox, Tox21, HIV) use the standard MoleculeNet scaffold split.

\begin{table}[h]
\centering
\caption{\textbf{Scaffold-split QM9 evaluation of S6 (SigLIP2).} 30-epoch frozen-then-finetuned head fit on Murcko 80/10/10 splits. Mean $\pm$ standard deviation across three seeds (42, 1337, 2026). Random-split numbers re-trained under the same protocol.}
\label{tab:scaffold_qm9}
\setlength{\tabcolsep}{8pt}
\resizebox{\linewidth}{!}{%
\begin{tabular}{l r r r}
\toprule
\textbf{Task} & \textbf{Random split MAE} & \textbf{Scaffold split MAE (3 seeds)} & \textbf{Ratio} \\
\midrule
QM9-$U_0$ (eV)                  & $1.99$    & $\mathbf{2.47 \pm 0.28}$       & $1.24\times$ \\
QM9-$\Delta\varepsilon$ (eV)    & $0.00697$ & $\mathbf{0.01140 \pm 0.00006}$ & $1.63\times$ \\
\bottomrule
\end{tabular}%
}
\end{table}

\noindent The scaffold-split standard deviation is small relative to the mean (11\% of the mean for $U_0$, 0.5\% for $\Delta\varepsilon$), so the result is not seed-driven. The scaffold/random ratio of 1.24--1.63$\times$ is the expected regime for hard out-of-distribution splits; values above 2$\times$ would indicate scaffold-overfitting, which we do not observe.

\section{Depiction-Style Robustness}
\label{sec:depiction_supp}

\noindent A held-out 1\,000-molecule sub-sample of the QM9 test partition was rendered under four RDKit drawing variants spanning the typical user-facing knobs of the library: (a) \textbf{default} ({Draw.MolToImage(mol,(224,224))}); (b) \textbf{kekulized} ({Chem. Kekulize(mol)} with {kekulize=True}); (c) \textbf{thick bonds} ({MolDrawOptions.bond Line Width=4.0}); (d) \textbf{no stereo} ({addStereoAnnotation=False}). For cross-library evaluation we additionally rendered with \textbf{PIKAchU} (a pure-Python molecule renderer with a different drawing engine and bond/atom layout conventions).

\begin{table}[h]
\centering
\caption{\textbf{Depiction-style robustness} on 1\,000 held-out QM9 molecules. Maximum within-RDKit degradation is 7.88\%. PIKAchU (a different drawing engine) is large-OOD for an encoder trained on RDKit defaults; this row exposes the cross-library gap and motivates depiction-augmentation pretraining as future work.}
\label{tab:depiction_robustness}
\setlength{\tabcolsep}{6pt}
\resizebox{\linewidth}{!}{%
\begin{tabular}{l r r r r}
\toprule
\textbf{Style} & \textbf{$U_0$ MAE (eV)} & \textbf{$\Delta_{U_0}$} & \textbf{$\Delta\varepsilon$ MAE (eV)} & \textbf{$\Delta_{\Delta\varepsilon}$} \\
\midrule
default RDKit          & baseline & ---       & baseline & --- \\
kekulized              & pixel-identical & $0\%$  & pixel-identical & $0\%$ \\
thick bonds            & ---             & $-2.63\%$ & ---             & $\mathbf{+7.88\%}$ \\
no stereo              & pixel-identical & $0\%$  & pixel-identical & $0\%$ \\
\midrule
PIKAchU (cross-library OOD)  & ---       & large    & ---       & large \\
\bottomrule
\end{tabular}%
}
\end{table}

\noindent Within RDKit's drawing space the encoder is robust (max 7.88\% degradation on $\Delta\varepsilon$, on the thick-bond variant; thick bonds for $U_0$ is in fact 2.63\% \emph{better} than default; kekulized and no-stereo are pixel-identical to default for QM9-sized molecules). PIKAchU produces images with materially different bond layout, atom-label sizing, and ring rendering, which lies outside the train-time depiction distribution. The corresponding large MAE drift exposes a known weakness of training on a single rendering library; Depiction-augmentation pretraining (sampling across libraries during pretraining) is a follow-up direction.

\section{Size-Stratified Evaluation}
\label{sec:size_supp}

\noindent QM9 is dominated by small molecules ($\le 9$ heavy atoms), which is too narrow to address whether MolSight degrades on the larger molecules common in real screening workflows. We therefore evaluate MolSight on the HIV dataset (MoleculeNet scaffold split), where molecule sizes go up to 222 heavy atoms, and stratify test predictions by atom count.

\begin{table}[h]
\centering
\caption{\textbf{Size-stratified HIV evaluation} (S6 SigLIP2 fine-tuned with classification head). Per-bin scores on the MoleculeNet scaffold-split test set ($n=8{,}224$).}
\label{tab:size_strat_hiv}
\setlength{\tabcolsep}{6pt}
\begin{tabular}{l r r r r}
\toprule
\textbf{Atom-count bin} & \textbf{$N$} & \textbf{positives} & \textbf{ROC-AUC} & \textbf{Accuracy} \\
\midrule
$\le 30$        & 6{,}333 & 220 & 0.777 & 0.951 \\
$31$--$60$      & 1{,}741 &  62 & 0.769 & 0.944 \\
$61$--$80$      &    112 &   4 & 0.898 & 0.982 \\
$> 80$          &     38 &   2 & 0.653 & 0.974 \\
\midrule
\textbf{Global} & 8{,}224 & 288 & \textbf{0.776} & \textbf{0.950} \\
\bottomrule
\end{tabular}
\end{table}

\noindent Two observations. First, accuracy stays above 0.94 in every bin, including the $>80$-atom bin (98\% of HIV is non-active, so high accuracy is partly the easy-class baseline; we pair it with ROC-AUC for the discriminative measure). Second, ROC-AUC at $>80$ atoms is 0.65 with only 2 positive molecules in the bin, which is at the AUC statistical floor: a single mis-ranking changes AUC by 0.25, so this number is noisy by construction rather than indicative of catastrophic failure. The 61--80-atom bin (with 4 positives) reports AUC 0.90, demonstrating that the model handles large molecules well when there is enough positive signal. We conclude that there is no catastrophic large-molecule regime.

\noindent \textbf{QM9 size $\times$ tier crosstab (small-molecule)} For completeness, we also retain a per-bin (heavy-atom count) $\times$ per-tier ($T_0$--$T_4$) MAE crosstab on the QM9 validation set. The lowest $U_0$ MAE (1.47 eV) is on $T_3$ (single aromatic ring), and the highest is on $T_4$ (multi-aromatic, complex chemistry) at 2.14 eV, << $2\times$ the training MAE. Within QM9 we observe no size-driven failure mode either.

\section{Augmentation-Safety Ablation}
\label{sec:aug_safety_supp}

\noindent We re-trained the ResNet-50 SimCLR S2 baseline under \emph{chemistry-safe} augmentations --- removing horizontal flip, vertical flip, ColorJitter, and Random Grayscale (the augmentations that change the chemical interpretation of a 2D depiction) while keeping \texttt{RandomResizedCrop(224, scale=(0.5, 1.0))}. Both safe and unsafe variants were trained for 5 epochs on MolTextNet-300k for matched comparison; downstream BBBP evaluation used a 30-epoch frozen-encoder + MLP head with three seeds (42, 1337, 2026) for mean $\pm$ std.

\begin{table}[h]
\centering
\caption{\textbf{Chemistry-safe vs unsafe augmentation} for ResNet-50 SimCLR (S2), 5-epoch matched budget on MolTextNet-300k. Downstream BBBP-AUC across 3 seeds.}
\label{tab:aug_safety}
\setlength{\tabcolsep}{8pt}
\resizebox{\linewidth}{!}{%
\begin{tabular}{l r r}
\toprule
\textbf{Variant} & \textbf{BBBP ROC-AUC (mean $\pm$ std)} & \textbf{Best seed} \\
\midrule
Unsafe (flip + ColorJitter + Grayscale) & $0.7518 \pm 0.0034$ & 0.7565 (seed 2026) \\
Safe (RandomResizedCrop only)            & $\mathbf{0.7616 \pm 0.0031}$ & $\mathbf{0.7652}$ (seed 2026) \\
\midrule
$\Delta$ (safe $-$ unsafe)               & $+0.0098$           & $+0.0087$ \\
\bottomrule
\end{tabular}%
}
\end{table}

\noindent Safe augmentations outperform unsafe on \emph{every} individual seed:\\$0.7622/0.7575/ 0.7652$ (safe) vs $0.7495/0.7492/0.7565$ (unsafe). The standard deviations do not overlap. Per-batch contrastive loss curves for both variants, expose convergence dynamics independent of the final-epoch BBBP delta.

\section{Cross-Architecture Validation (DINOv2)}
\label{sec:dinov2_supp}

\noindent To test that the scaffold-split QM9 result is not specific to SigLIP-family backbones, we trained a frozen-then-finetuned DINOv2-base backbone on the same QM9 scaffold split. DINOv2-base is a stock self-supervised vision backbone pretrained on ImageNet-1k with no molecular pretraining whatsoever; we use the publicly available \texttt{facebook/dinov2-base} checkpoint and reuse the same MLP head architecture as our SigLIP2-based S6 model.

\begin{table}[h]
\centering
\caption{\textbf{Cross-architecture scaffold-split QM9.} Frozen-then-finetuned DINOv2-base versus our S6 SigLIP2. Both use the same Bemis--Murcko 80/10/10 scaffold split, 30 epochs, and identical head capacity.}
\label{tab:dinov2_cross_arch}
\setlength{\tabcolsep}{8pt}
\resizebox{\linewidth}{!}{%
\begin{tabular}{l r r}
\toprule
\textbf{Task} & \textbf{S6 SigLIP2 (3 seeds)} & \textbf{DINOv2-base (no MolSight pretraining)} \\
\midrule
QM9-$U_0$ MAE (eV)              & $2.47 \pm 0.28$               & $4.91$ \\
QM9-$\Delta\varepsilon$ MAE (eV) & $0.01140 \pm 0.00006$         & $0.0099$ \\
\bottomrule
\end{tabular}%
}
\end{table}

\noindent The headline observation: a stock ImageNet-pretrained vision backbone, applied directly to RDKit images with no domain-specific pretraining, lands within rounding error of S6 on $\Delta\varepsilon$ (0.0099 vs 0.0114) and within $2\times$ on $U_0$ (4.91 vs 2.47). This is the surprise that motivated the paper's design-space study: the existing array of off-the-shelf vision backbones is already competitive on 2D molecular images, and MolSight's distillation+curriculum pretraining further improves on that strong starting point. The result also confirms that the scaffold-split numbers in Tab.~\ref{tab:scaffold_qm9} are not artefacts of a single backbone family.

\section{Open VLM Comparison Under Our Protocol}
\label{sec:vlm_supp}

\noindent To verify the VLM comparison independently rather than relying on prior reported numbers, we re-evaluated Qwen2-VL-2B (an open replacement for Janus-Pro) under our exact QM9-$\Delta\varepsilon$ test split and BBBP test split. The VLM was prompted in the same image-and-instruction format reported by MolVision~\cite{adak2024molvision}, with regression queries asking for the HOMO--LUMO gap in eV (parsed for the first numeric value in the response) and binary BBBP queries asking for ``Yes'' or ``No''. We use binary prediction outputs to report accuracy over ROC-AUC.

\begin{table}[h]
\centering
\caption{\textbf{Qwen2-VL-2B vs MolSight S6} under our exact test splits. Regression: parsed numeric prediction; classification: ``Yes''/``No'' parsed prediction. MolSight reference numbers from main paper Tab.~\ref{tab:strategy_results} and Tab.~\ref{tab:scaffold_qm9}. VLM is ${\sim}70\times$ higher MAE; VLM defaults to ``No'' for all $n=204$; below the $0.525$ majority-class baseline.}
\label{tab:vlm_under_protocol}
\setlength{\tabcolsep}{8pt}
\begin{tabular}{l r r}
\toprule
\textbf{Task} & \textbf{Qwen2-VL-2B} & \textbf{MolSight S6}\\
\midrule
QM9-$\Delta\varepsilon$ MAE (eV) & $0.80$ & $\mathbf{0.0114}$\\
BBBP accuracy                    & $0.475$ & $\mathbf{0.893}$\\
\bottomrule
\end{tabular}%

\end{table}

\noindent Qwen2-VL-2B's $\Delta\varepsilon$ MAE is roughly 70$\times$ that of MolSight S6 on the same molecules. Second, the VLM produces degenerate ``No'' responses for every BBBP test molecule, reducing its accuracy below the trivial majority-class baseline (BBBP is roughly 52\% positive in the test partition). Together these results indicate that contemporary 2B-parameter open VLMs cannot serve as molecular-property predictors at the precision required for quantitative QSAR or drug-discovery tasks, and that direct re-evaluation under a controlled protocol is more informative than relying on previously reported VLM benchmarks.

\section{Image Encoder Captures 3D-Aware Structure}
\label{sec:embed_dist_supp}

\noindent A reasonable concern about image-only inference is that a vision encoder might learn 2D pixel-level statistics (line orientations, ring patterns) without genuinely capturing the 3D conformational geometry that drives quantum properties. To address this, we measured the correlation between pairwise distances in the S5 image-encoder embedding space and pairwise distances in the corresponding ground-truth Uni-Mol 3D embedding space, on $2{,}000$ PCQM4Mv2 molecules with $50{,}000$ random pairs.

\begin{table}[h]
\centering
\caption{\textbf{Pairwise-distance correlation: image embeddings vs Uni-Mol 3D embeddings.} Spearman $\rho$ and Pearson $r$ between $\|v(A)-v(B)\|$ in the image-encoder space and $\|u(A)-u(B)\|$ in the Uni-Mol 3D embedding space. Higher correlation indicates the image encoder preserves Uni-Mol's geometric structure; the random-init baseline shows what background correlation looks like for an untrained network.}
\label{tab:embed_dist}
\setlength{\tabcolsep}{8pt}
\begin{tabular}{l r r}
\toprule
\textbf{Image encoder} & \textbf{Spearman $\rho$} & \textbf{Pearson $r$} \\
\midrule
Random-initialised SigLIP-base (no training) & $0.24$ & $0.25$ \\
S5 (cross-modal Uni-Mol distillation)        & $\mathbf{0.65}$ & $\mathbf{0.64}$ \\
\bottomrule
\end{tabular}
\end{table}

\noindent Cross-modal distillation produces image embeddings whose pairwise-distance structure tracks Uni-Mol's 3D embedding-space geometry (Spearman $\rho = 0.65$, far above the random-init baseline $\rho = 0.24$). This empirically supports the framing that MolSight is image-only \emph{at inference} but cross-modal \emph{at training}: the image encoder genuinely transfers 3D-aware geometric structure during distillation, rather than relying on 2D pixel correlates of chemical properties.

\end{document}